\documentclass[letterpaper]{article} 
\usepackage{aaai24}  
\usepackage{times}  
\usepackage{helvet}  
\usepackage{courier}  
\usepackage[hyphens]{url}  
\usepackage{graphicx} 
\usepackage{natbib}  
\usepackage{caption} 
\usepackage{algorithm}
\usepackage{algorithmic}

\usepackage{epsfig}
\usepackage{graphicx}
\usepackage{amsthm}
\usepackage{amsmath}
\usepackage{amssymb}
\usepackage{booktabs}
\usepackage{algorithm}
\usepackage{algorithmic}
\usepackage{bbding}
\usepackage{subcaption}
\usepackage{multirow}
\usepackage{bm}
\usepackage{xr}
\usepackage{xspace}

\usepackage{newfloat}
\usepackage{listings}
\DeclareCaptionStyle{ruled}{labelfont=normalfont,labelsep=colon,strut=off} 
\floatstyle{ruled}
\newfloat{listing}{tb}{lst}{}
\floatname{listing}{Listing}

\usepackage{caption}
\usepackage[dvipsnames]{xcolor}  
\usepackage{listings}
\makeatletter
\DeclareRobustCommand\onedot{\futurelet\@let@token\@onedot}
\def\@onedot{\ifx\@let@token.\else.\null\fi\xspace}

\newcommand{\vect}[1]{\boldsymbol{\mathbf{#1}}}
 
\def\ie{\emph{i.e}\onedot}

\makeatother

\newtheorem{theorem}{Theorem}[section]

\newtheorem{lemma}[theorem]{Lemma}
\theoremstyle{definition}

\title{Make RepVGG Greater Again: A Quantization-aware Approach}
\author{
	Xiangxiang Chu, Liang Li, Bo Zhang\\
}
\affiliations{
	chuxiangxiang, liliang58, zhangbo97@meituan.com \\
	Meituan\\
	
}

\usepackage{bibentry}
\makeatletter
\newcommand*{\addFileDependency}[1]{
	\typeout{(#1)}
	\@addtofilelist{#1}
	\IfFileExists{#1}{}{\typeout{No file #1.}}
}
\makeatother
\newcommand*{\myexternaldocument}[1]{%
	\externaldocument{#1}%
	\addFileDependency{#1.tex}%
	\addFileDependency{#1.aux}%
}
\myexternaldocument{qarep_aaai24_supp}

\begin{document}

\maketitle

\begin{abstract}
	The tradeoff between performance and inference speed is critical for practical applications. Architecture reparameterization obtains better tradeoffs and it is becoming an increasingly popular ingredient in modern convolutional neural networks. Nonetheless, its quantization performance is usually too poor to deploy (more than 20\% top-1 accuracy drop on ImageNet) when INT8 inference is desired. In this paper, we dive into the underlying mechanism of this  failure, where the original design inevitably enlarges quantization error.  We propose a simple, robust,  and effective remedy to have a quantization-friendly structure that also enjoys reparameterization benefits. Our method greatly bridges the gap between INT8 and FP32 accuracy for RepVGG. Without bells and whistles,  the top-1 accuracy drop on ImageNet is reduced within 2\% by  standard  post-training  quantization. Moreover, our method also achieves similar FP32 performance as RepVGG.  Extensive experiments on  detection and  semantic segmentation tasks verify its generalization.   
\end{abstract}

\section{Introduction}\label{sec:intro}

Albeit the great success of deep neural networks in vision \cite{he2016deep, he2017mask, chen2017deeplab,redmon2016you, dosovitskiy2020image}, language \cite{vaswani2017attention, devlin2019bert} and speech \cite{graves2013speech}, \emph{model compression} has become more than necessary, especially considering the paramount growth of power consumption in data centers, and the voluminous distribution of resource-constrained edge devices worldwide. Network quantization \cite{gupta2015deep,gysel2018ristretto} is one the most proficient approaches because of the lower memory cost and inherent integer computing advantage.

Still, quantization awareness in neural architectural design has not been the priority and has thus been largely neglected. However, it may become detrimental where quantization is a mandatory operation for final deployment. For example, many well-known architectures have quantization collapse issues like MobileNet \cite{howard2017mobilenets,sandler2018mobilenetv2,howard2019searching} and EfficientNet \cite{tan2019efficientnet}, which calls for remedy designs or advanced quantization schemes like \cite{sheng2018quantization,yun2021all} and \cite{bhalgat2020lsq+,habi2021hptq} respectively.

\begin{figure}[ht]
	\centering
	\includegraphics[width=0.75\columnwidth]{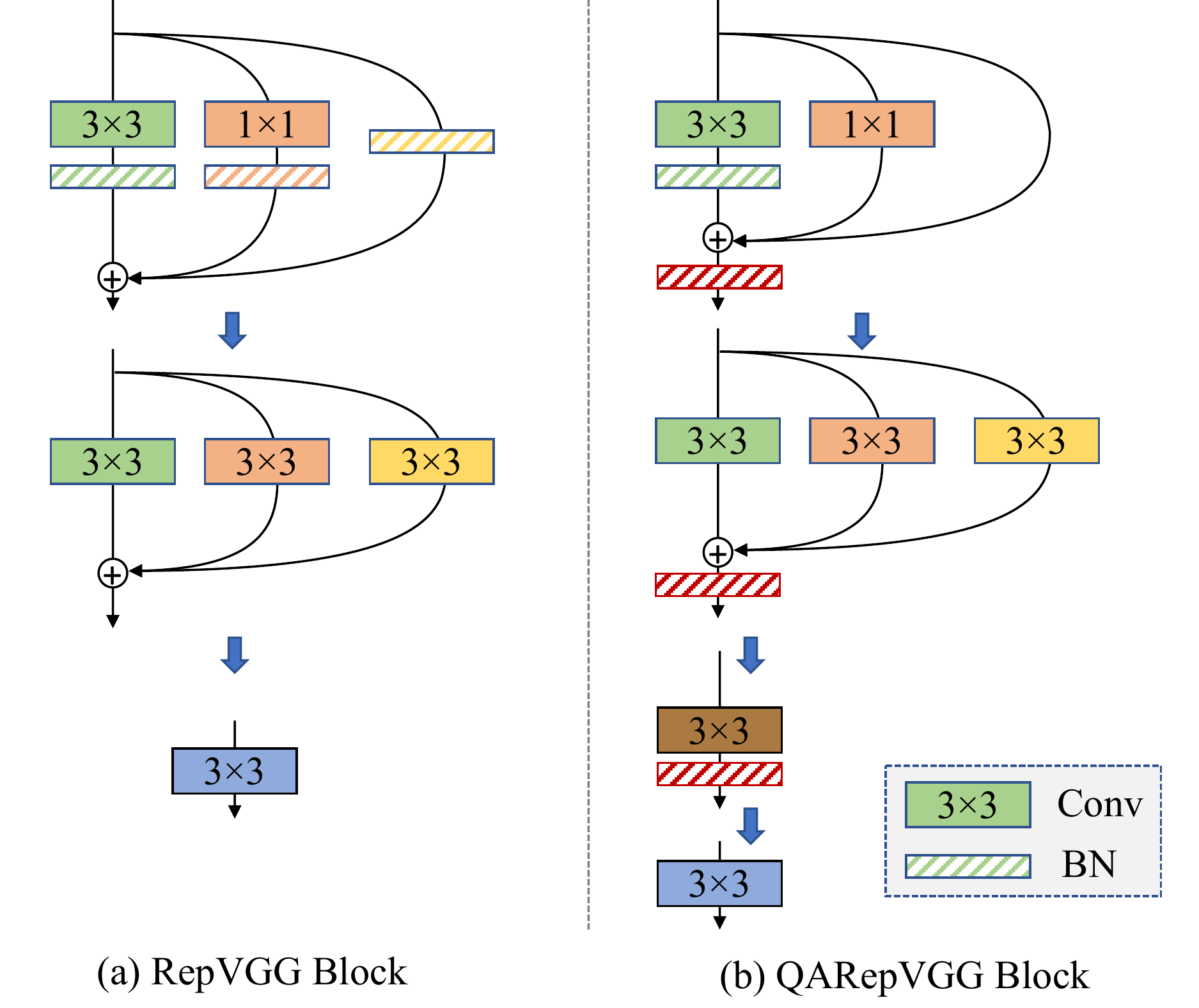}
	\caption{Reparameterization of a QARepVGG block compared with RepVGG \cite{ding2021repvgg}. Both can be identically fused into a single Conv $3\times3$. QARepVGG is PTQ-friendly to have \textbf{70.4\%} quantized accuracy while RepVGG drops to \textbf{52.2\%}.}
	\label{fig:qarepvgg-block}
\end{figure}

Lately, one of the most influential directions in neural architecture design has been reparameterization \cite{zagoruyko2017diracnets,ding2019acnet,ding2021repvgg}. Among them, RepVGG \cite{ding2021repvgg} refashions the standard Conv-BN-ReLU into its identical multi-branch counterpart during training, which brings powerful performance improvement while adding no extra cost at inference. For its simplicity and inference advantage, it is favored by many recent vision tasks \cite{ding2022scaling,xu2022pp,li2022yolov6,wang2022yolov7,vasu2022improved,hu2022online}. However, reparameterization-based models face a well-known \emph{quantization difficulty} which is an intrinsic defect that stalls industry application. It turns out to be non-trivial to make this structure comfortably quantized. A standard post-training quantization scheme tremendously degrades the accuracy of RepVGG-A0 from 72.4\% to 52.2\%. Meantime, it is not straightforward to apply quantization-aware training \cite{ding2023reparameterizing}.

Here, we particularly focus on the quantization difficulty of RepVGG \cite{ding2021repvgg}. To resolve this problem, we explore the fundamental quantization principles that guide us through in-depth analysis of the typical reparameterization-based architecture. That is, for a network to have better quantization performance, the distribution of weights as well as the processed data of an arbitrary distribution shall be quantization friendly. Both are crucial to ensure better quantization performance. More importantly, these principles lead us to a brand new design which we call QARepVGG (short for Quantization-Aware RepVGG) that doesn't suffer from substantial quantization collapse, whose building block is shown in Fig.~\ref{fig:qarepvgg-block} and its quantized performance has been largely improved.


Our contributions are threefold,
\begin{itemize}
	\itemsep0em 
	\item Unveiling the root cause of performance collapse in the  quantization of the reparameterization-based architecture like RepVGG.
	\item Contriving a quantization-friendly replacement (\ie QARepVGG) of RepVGG which holds fundamental differences in terms of weight and activation distribution, meanwhile preserving the very advantage of outstanding speed and performance trade-off.
	\item Our proposed method generalizes well at different model scales and on various vision tasks, achieving outstanding post-quantization performance that is ready to deploy. Besides, our method has  comparable FP32 accuracy  as RepVGG and exactly the same fast inference speed under the deploy setting. Therefore, it is a very competitive alternative to RepVGG. 
\end{itemize}

Expectedly, our approach will greatly boost the quantized performance with no extra cost at inference, bridging the gap of the last kilometer during the deployment of reparamenterized networks. We will release the code to facilitate reproduction and future research.

\section{Related Work}
\textbf{Reparameterization Architecture Design.} RepVGG \cite{ding2021repvgg} leverages an over-parameterized network in the form of multiple branches at the training stage and identically fuses branches into one during inference, which is known as reparameterization. This design is becoming wildly used as a basic component in many scenarios, such as edge device application  \cite{vasu2022improved,zhou2023fastpillars,pmlr-v162-wu22f,huang2022rife} , high performance convnet \cite{ding2022scaling,huang2022dyrep}, covering both low-level and high-level vision tasks.
Recent popular object detection methods like  YOLOv6 \cite{li2022yolov6} and YOLOv7 \cite{wang2022yolov7} are both built based on such basic component.   OREPA \cite{hu2022online} is  a structural improvement on RepVGG, which aims to
reduce the huge training overhead by squeezing the complex
training-time block into a single convolution. However, almost all these researches make use of the high FP32 performance of reparameterization and fast inference under the deploy setting.

\textbf{Quantization}. \emph{Network Quantization} is an effective model compression method that maps the network weights and input data into lower precisions (typically 8-bit) for fast calculations, which greatly reduces the model size and computation cost. Without compromising much  performance, quantization is mostly adopted to boost speed before deployment, serving as a de facto standard in industrial production. Post-Training Quantization (PTQ) is the most common scheme as it only needs a few batches of images to calibrate the quantization parameters and it comes with no extra training. Quantization-Aware Training (QAT) methods have also been proposed to improve the quantized accuracy, such as integer-arithmetic-only quantization \cite{jacob2018quantization}, data-free quantization \cite{nagel2019data}, hardware-aware quantization \cite{wang2019haq}, mixed precision quantization \cite{wu2018mixed}, zero-shot quantization \cite{cai2020zeroq}. As QAT typically involves intrusion into the training code and requires extra cost, it is only used when the training code is at hand and PTQ can't produce a satisfactory result. To best showcase the proposed quantization-aware architecture, we mainly evaluate the quantized accuracy using PTQ. Meanwhile, we include experiments to demonstrate it is also beneficial for QAT.


\textbf{Quantization for Reparameterization Network}. It is known that reparameterization-based architectures have quantization difficulty due to the increased dynamic numerical range due to its intrinsic multi-branch design  \cite{ding2023reparameterizing}.  The accuracy degradation of reparameterization models via PTQ is unacceptable. The most related work to ours is
RepOpt-VGG, which makes an attempt to address this quantization issue by crafting a two-stage optimization pipeline.
However, it requires very careful hyper-parameter tuning to work and more computations. In contrast, our method is neat, robust and computation cheap.


\section{Make Reparameterization Quantization Friendly}
This section is organized as follows. First, we disclose that the popular reparameterization design of RepVGG models severely suffers from quantization. Then we   make detailed analysis of the root causes for the failure and reveal that two factors incurs this issue: the loss design enlarges the variance of activation and the structural design  of RepVGG is prone to producing uncontrolled outlier  weights.  Lastly, we greatly alleviate  the quantization issue  by revisiting loss and network design.




\subsection{Quantization Failure of  RepVGG}
We first evaluate  the performance of several RepVGG models with its officially released code.
As shown in Table~\ref{tab:abation_repvgg}, RepVGG-A0 serevely  suffers from large accuracy drop (from 20\% to 77\%  top-1 accuracy) on ImageNet validation data-set after standard PTQ.

A quantization operation for a tensor $X$ is generally represented as $Q(X) = Clip(Round(X/\bigtriangleup_{x}))$, where $Round$ rounds float values to integers using ceiling rounding and $Clip$ truncates those exceed the ranges of the quantized domain. $\bigtriangleup_{x}$ is a scale factor used to map the tensor into a given range, defined as $\bigtriangleup _{x} = \frac{x_{max}-x_{min}  }{2^{b}-1 } $. Where $x_{max}$ and $x_{min}$ are a pair of boundary values selected to better represent values distribution of $X$. As shown in \cite{dehner2016analysis} and \cite{sheng2018quantization}, the variance of the quantization error is calculated as $\sigma ^{2}  =\frac{\bigtriangleup _{x}^{2} }{12} $. Thus the problem becomes how to reduce the range between $x_{max}$ and $x_{min}$. In practice, they are selected in various ways. Sometimes the maximum and minimum values are used directly, such as weight quantization, and sometimes they are selected by optimizing the MSE or entropy of the quantization error, which is often used to quantify the activation value. The quality of the selection depends on many factors, such as the variance of the tensor, whether there are some outliers, etc.

%
As for a neural network, there are two main components, weight and activation, that require quantization and may lead to accuracy degradation. Activation also serves as the input of the next layer, so the errors are accumulated and incremented layer by layer. Therefore, good quantization performance for neural networks requires mainly two fundamental conditions:
\begin{itemize}
	\item [\textbf{C1}] weight distribution is  quantization friendly with feasible range,
	\item[\textbf{C2}] activation distribution (\ie how the model responds to input features) is also friendly for quantization.
\end{itemize}

Empirically, we define a distribution of weights or activations as \textit{quantization friendly} if it has a small variance and few outliers. Violating either one of above conditions will lead to inferior quantization performance. 

We use RepVGG-A0 as an example to study why the quantization of the reparameterization-based structure is difficult. We first reproduce the performance of RepVGG with its officially released code, shown in Table~\ref{tab:abation_repvgg}. Based on this, we can further strictly control the experiment settings. We quantize RepVGG-A0 with a standard setting of PTQ and evaluate the INT8 accuracy, which is dropped from 72.2\% to 50.3$\%$. Note that we use the deployed model after fusing multi-branches, because the unfused one would incur extra quantization errors.  This trick is widely used in popular quantization frameworks.
\begin{table}[h]
	\setlength{\tabcolsep}{2pt}  
	\centering
	\begin{tabular}{lclc|}
		\toprule
		Variants			&FP32 Acc 	& INT8 Acc	\\
		&   (\%) & (\%) \\
		\midrule
		RepVGG-A0 (w/ custom $L_2$)$^\star$ & 72.4& 52.2 (20.2$\downarrow$)\\
		RepVGG-A0 (w/ custom $L_2$)$^\dagger$ & 72.2& 50.3 (21.9$\downarrow$)\\
		\bottomrule
	\end{tabular}
	
	\caption{Quantizied top-1 accuracy on ImageNet using RepVGG-A0.  $^\star$: from the official repo. $^\dagger$: reproduced with the official code. }
	\label{tab:abation_repvgg}
	\vskip -0.2in
\end{table}

We illustrate the weight distribution of our reproduced model RepVGG-A0 in Fig.~\ref{fig:repvgg-a0-boxplot} and Fig.~\ref{fig:repvgg-a0-weight-dist} (appendix). Observing that the weights are well distributed around zero and no particular outlier exists, it satisfies \textbf{C1}. This leads us to verify \textbf{C2} if it is the activation that greatly deteriorates the quantization.  Unfortunately, the activation is input-dependent and coupled with the learned weights. We hereby don't impose any assumptions on the distribution of weight or activation and 
analyze the deviation of each branch. 

\subsubsection{Regularized loss  enlarges the  activation variance}
Before we proceed, we formulate the computation operations in a typical RepVGG block. We  keep  the same naming convention as \cite{ding2021repvgg} to be better understood. Specifically, we use $\mathrm{W}_{(k)}\in\mathbb{R}^{C_2\times C_1\times k\times k}$ to denote the kernel of a $k\times k$ convolution, where  $C_1$  and $C_2$  are the number of input  and  output channels respectively. Note that  $k \in \{1,3\}$ for RepVGG. As for the batch normalization (BN) layer after $k\times k$ convolution, we use $\vect{\mu}_{(k)}\in \mathbb{R}^{C_2},\vect{\sigma}_{(k)}\in \mathbb{R}^{C_2},\vect{\gamma}_{(k)}\in \mathbb{R}^{C_2},\vect{\beta}_{(k)}\in \mathbb{R}^{C_2}$ as the mean, standard deviation, scaling factor and the bias.  For the BN in the identity branch, we use  $\vect{\mu}_{(0)},\vect{\sigma}_{(0)},\vect{\gamma}_{(0)},\vect{\beta}_{(0)}$. Let $\mathrm{M}_{(\text{1})}\in\mathbb{R}^{N\times C_1\times H_1\times W_1}$, $\mathrm{M}_{(\text{2})}\in\mathbb{R}^{N\times C_2\times H_2\times W_2}$ be the input and output respectively, and `$\ast$' be the convolution operator. Let Y$_{(\text{0})}$,Y$_{(\text{1})}$ and Y$_{(\text{3})}$ be the output of the Identity, 1$\times$1 and 3$\times$3 branch.  Without loss of generality, we suppose   $C_1=C_2, H_1=H_2, W_1=W_2$. Then we can write the output $\mathrm{M}_{(\text{2})}$ as,


\begin{equation}
	\label{eq:bn_add}
	\begin{aligned}
		\mathrm{M}_{(\text{2})} 
		&= \text{BN}(\mathrm{M}_{(\text{1})} \ast \mathrm{W}_{(3)},\vect{\mu}_{(3)},\vect{\sigma}_{(3)},\vect{\gamma}_{(3)},\vect{\beta}_{(3)}) \\
		&\quad+\text{BN}(\mathrm{M}_{(\text{1})} \ast \mathrm{W}_{(1)},\vect{\mu}_{(1)},\vect{\sigma}_{(1)},\vect{\gamma}_{(1)},\vect{\beta}_{(1)}) \\
		&\quad+\text{BN}(\mathrm{M}_{(\text{1})},\vect{\mu}_{(0)},\vect{\sigma}_{(0)},\vect{\gamma}_{(0)},\vect{\beta}_{(0)}) \,.\\		
	\end{aligned}
\end{equation}

The batch normalization operation for the  3$\times$3 branch can be written as,
\begin{equation}
	\label{eq:conv3}
	\mathrm{Y}_{(3)} = \vect{\gamma}_{(3)}\odot\frac{\mathrm{M}_{(1)}\ast  \mathrm{W}_{(3)}-\vect{\mu}_{(\mathrm{3})}}{\sqrt{\epsilon+\vect{\sigma}_{(3)}\odot\vect{\sigma}_{(3)}}}+\vect{\beta}_{(3)},
\end{equation}
where $\odot$ is element-wise multiplication and  $\epsilon$ a small value (10$^{-5}$ by default) for numerical stability. This means BN plays a role of changing the statistic (mean and variance) of its input.  \textit{Note that the change of $\vect{\mu}$ doesn't  affect the quantization error. However, the changed variance directly affects the quantization accuracy.}
From the probability perspective, given a random variable X,  and a scalar $\lambda$, the variance of $\lambda X$, i.e.  $D(\lambda$X) equals to  $\lambda^2D$(X). Let $X_{(3)} =\mathrm{M}^{(1)}\mathrm{W}_{(3)} $,  then we have   

\begin{equation}
	D(\mathrm{Y}_{(3)})=\frac{\vect{\gamma}_{(3)}\odot\vect{\gamma}_{(3)}}{{\epsilon+\vect{\sigma}_{(3)}\odot\vect{\sigma}_{(3)}}}\odot D(X_{(3)}).
	\label{eq:var_exp}
\end{equation}
The value of  $\frac{\vect{\gamma}_{(3)}\odot\vect{\gamma}_{(3)}}{{\epsilon+\vect{\sigma}_{(3)}\odot\vect{\sigma}_{(3)}}}$  controls shrinking or expanding the variance of $X_{(3)}$, which in turn leads to better or worse quantization performance respectively. For 1$\times$1 and Identity, we can draw similar conclusions.

\begin{figure}[ht]
	\centering
	\includegraphics[width=0.85\columnwidth]{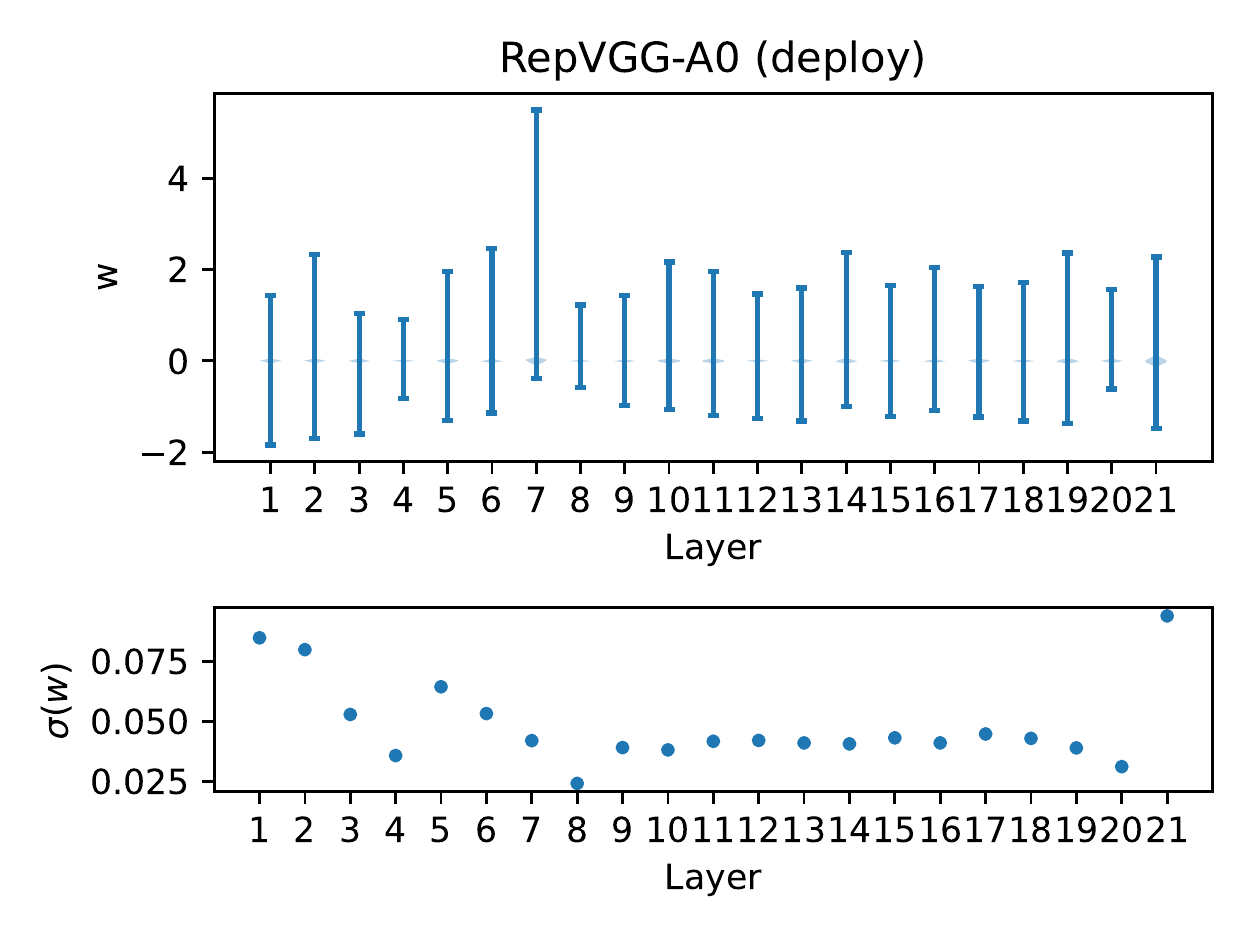}
	\caption{Violin plot and standard deviation of RepVGG-A0 convolutional weights per layer.The designed customed $L_2$ helps learned weights quantization friendly.}
	\label{fig:repvgg-a0-boxplot}
	\vskip -0.1in
\end{figure}

Based on the above analysis, we dive into the detail of RepVGG. There is a \emph{critical but easily neglected component},  which is a special design for the weight decay called custom $L_2$. It is stated that this component improves the accuracy and facilitates quantization  \cite{ding2021repvgg}. The detailed implementation is shown in Algorithm~\ref{alg:customL2} (appendix).  
This particular design regularizes the multi-branch weights as if it regularizes its equivalently fused kernel. It is likely to make the fused weights enjoy a quantization-friendly distribution\footnote{It can be verified by Fig.~\ref{fig:repvgg-a0-weight-dist} (appendix).  }. We illustrate the weight distribution of our reproduced model RepVGG-A0 in Fig.~\ref{fig:repvgg-a0-boxplot} and Fig.~\ref{fig:repvgg-a0-weight-dist} (appendix) and observe that the weights are well distributed around zero and no particular outlier exists.

This loss {\tt{l2\_loss\_eq\_kernel}} is essentially,
\begin{equation}
	L_{2_{custom}} = \frac{\lvert\mathrm{W}_{eq}\rvert_2^2}{\lvert \frac{\vect{\gamma_{(3)}}}{\sqrt{\epsilon+\vect{\sigma_{(3)}\odot\vect{\sigma_{(3)}}}}}\rvert^2_2+\lvert \frac{\vect{\gamma_{(1)}}}{\sqrt{\epsilon+\vect{\sigma_{(1)}}\odot\vect{\sigma_{(1)}}}}\rvert^2_2}.
	\label{eq:use_den}
\end{equation}

Notably, the optimizer are encouraged to \textbf{enlarge} the denominator  $\lvert \frac{\vect{\gamma_{(3)}}}{\sqrt{\epsilon+\vect{\sigma_{(3)}}\odot\vect{\sigma_{(3)}}}}\rvert^2_2+\lvert \frac{\vect{\gamma_{(1)}}}{\sqrt{\epsilon+\vect{\sigma_{(1)}\odot\vect{\sigma_{(1)}}}}}\rvert^2_2$ to minimize this loss, which \textbf{magnifies} the variance of activation and brings quantization difficulty.  This indicates that  custom $L_2$ helps to make learned weights quantization-friendly at the cost of activation  quantization-unfriendly. However,  we will show that  the model have troubles in quantizing learned weights without such regularized loss in the next section and this issue is inevitable because of the  structural design.

To address the variance enlarging issue, a simple and straight forward approach is   removing the denominator from Eq~\ref{eq:use_den} and we have
\begin{equation}
	L_{2_{custom}}' = \lvert\mathrm{W}_{eq}\rvert_2^2
	\label{eq:dis_den}
\end{equation}
We report the result in Table~\ref{tab:abation_dis_den}. Without the denominator item, the FP32 accuracy is 71.5\%, which is 0.7\% lower than the baseline. However, it's surprising to see that the quantization performance is greatly \textbf{improved} to 61.2\%.  However, this approach still requires inconvenient equivalent conversion.  

Another promising approach is applying  normal  $L_2$ directly.  Regarding, previous multi-branch networks like Inception series no longer require special treatment for weight decay, this motivates us to apply normal $L_2$ loss. The result is show in Table~\ref{tab:abation_dis_den}. Except for simplicity,  $L_2$ achieves slightly better result than Eq~\ref{eq:dis_den}. Therefore, we choose this approach as our default implementation (\textbf{M1}).




\begin{table}[ht]
	\setlength{\tabcolsep}{2pt}  
	\centering
	\begin{tabular}{lclc|}
		\toprule
		Variants			&FP32 Acc 	& INT8 Acc	\\
		&   (\%) & (\%) \\
		\midrule
		RepVGG-A0 (w/ custom $L_2$)$^\dagger$ & 72.2& 50.3 (21.9$\downarrow$)\\
		RepVGG-A0 (Eq~\ref{eq:dis_den})& 71.5& 61.2 (10.3$\downarrow$)\\
		RepVGG-A0 (w/ normal $L_2$)&71.7& 61.6 (10.1$\downarrow$)\\
		QARepVGG-A0 &72.2&\textbf{70.4} (1.8$\downarrow$)\\
		\bottomrule
	\end{tabular}
	\caption{Removing the denominator of custom $L_2$ improves the quantized top-1 accuracy on ImageNet. $^\dagger$: reproduced.}
	\label{tab:abation_dis_den}
\end{table}

%

\subsubsection{Structural design  of RepVGG is prone to producing uncontrolled outlier  weights}

While the FP32 accuracy is 0.5\% lower than the baseline, its INT8 accuracy is 11.3\% higher than the baseline. However, this design doesn't meet the application requirements either. Given that there are   no explicit regularizers to enlarge the activation variance, it is straightforward to check the distribution of weights. Firstly we can give the fused weight as,

\begin{equation}
	\small
	\begin{split}
		\mathrm{W} &= \mathrm{\hat{W}_{(3)}} + \mathrm{\hat{W}_{(1)}} + \mathrm{\hat{W}_{(0)}} \\
		&= \frac{ \vect{\gamma}_{(3)}}{\sqrt{\epsilon+\vect{\sigma}_{(3)}^{2} }} \odot\mathrm{W}_{(3)} + \frac{ \vect{\gamma}_{(1)}}{\sqrt{\epsilon+\vect{\sigma}_{(1)}^{2} }} \odot Padding(\mathrm{W}_{(1)}) \\
		&+ \frac{ \vect{\gamma}_{(0)}}{\sqrt{\epsilon+\vect{\sigma}_{(0)}^{2} }} \odot Padding( \mathrm{W}_{(0)} )
	\end{split}
\end{equation}

where $Padding()$ is applied to match shape of the 3$\times$3 kernel. In this architecture, $W_{(3)}$ and $W_{(1)}$ are trainable parameters, while $W_{(0)}$ is a fixed unit matrix that is not subject to decay during training. The scalars $\frac{ \vect{\gamma}_{(3)}}{\sqrt{\epsilon+\vect{\sigma}_{(3)}^{2} }}$ and $\frac{ \vect{\gamma}_{(1)}}{\sqrt{\epsilon+\vect{\sigma}_{(1)}^{2} }}$ depend on the outputs of the convolution layers of 3$\times$3 and 1$\times$1 branches, respectively. However, $\frac{\vect{\gamma}{(0)}}{\sqrt{\epsilon+\vect{\sigma}{(0)}^{2}}}$ directly depends on the output of the last layer. It is worth noting that the Identity branch is special because activations pass through a ReLU layer before entering a BatchNorm layer. This operation can be dangerous since if a single channel is completely unactivated (i.e., contains only zeros),  which generate a very small  $\sigma$ and a singular values of $\frac{1}{\sqrt{\epsilon+\vect{\sigma}_{(0)}^{2} }} $. This issue is common in networks that use ReLU widely. If this case occurs, the singular values will dominate the distribution of the fused kernels and significantly affect their quantization preference.

To verify our hypothesis, it is straightforward to check the distribution of weights, which is shown in Fig.~\ref{fig:repvgg-a0-nocwd-w-violin} (appendix) and Fig.~\ref{fig:repvgg-a0-nocwd-w-dist} (appendix). The \emph{fused weights distribution in both Layer 5 and 6 have large standard variances} (2.4 and 5.1 respectively), which are about two orders of magnitude larger than other layers. Specifically, the maximal values of fusedw eights from Layer 5 and 6 are 692.1107 and 1477.3740. This explains why the quantization performance is not good, violating \textbf{C1} causes unrepairable error. We further illustrate the $\frac{\gamma}{\sqrt{\epsilon+\sigma^2}}$ of three branches in Fig.~\ref{fig:repvgg-a0-nocwd-w-violin}. The maximal values of   $\frac{\gamma}{\sqrt{\epsilon+\sigma^2}}$ from the identity branch  on Layer 5 and 6 are 692.1107 and 1477.3732. It's interesting to see that the weights from the 3$\times$3 and 1$\times$ branches from Layer 4 also have some large values but their fused weights no longer contain  such values. 

We repeat the experiments thrice, and this phenomenon recurs. Note that the maximal values randomly occurs in different layers for different experiments. And simply skip those  layers could not solve the quantization problems. Partial quantization results of RepVGG-A0 (w/ custom $L_2$) in Table~\ref{tab:abation_dis_den} is only 51.3\%, after setting  Layer 5 and 6  float. According to our analysis, the quantization error for RepVGG is cumulated by all layers, so partial solution won't mitigate the collapse. This motivates us to address this issue by designing quantization-friendly reparameterization structures.

%
%
\begin{figure}[ht]
	\centering
	\includegraphics[width=0.95\columnwidth]{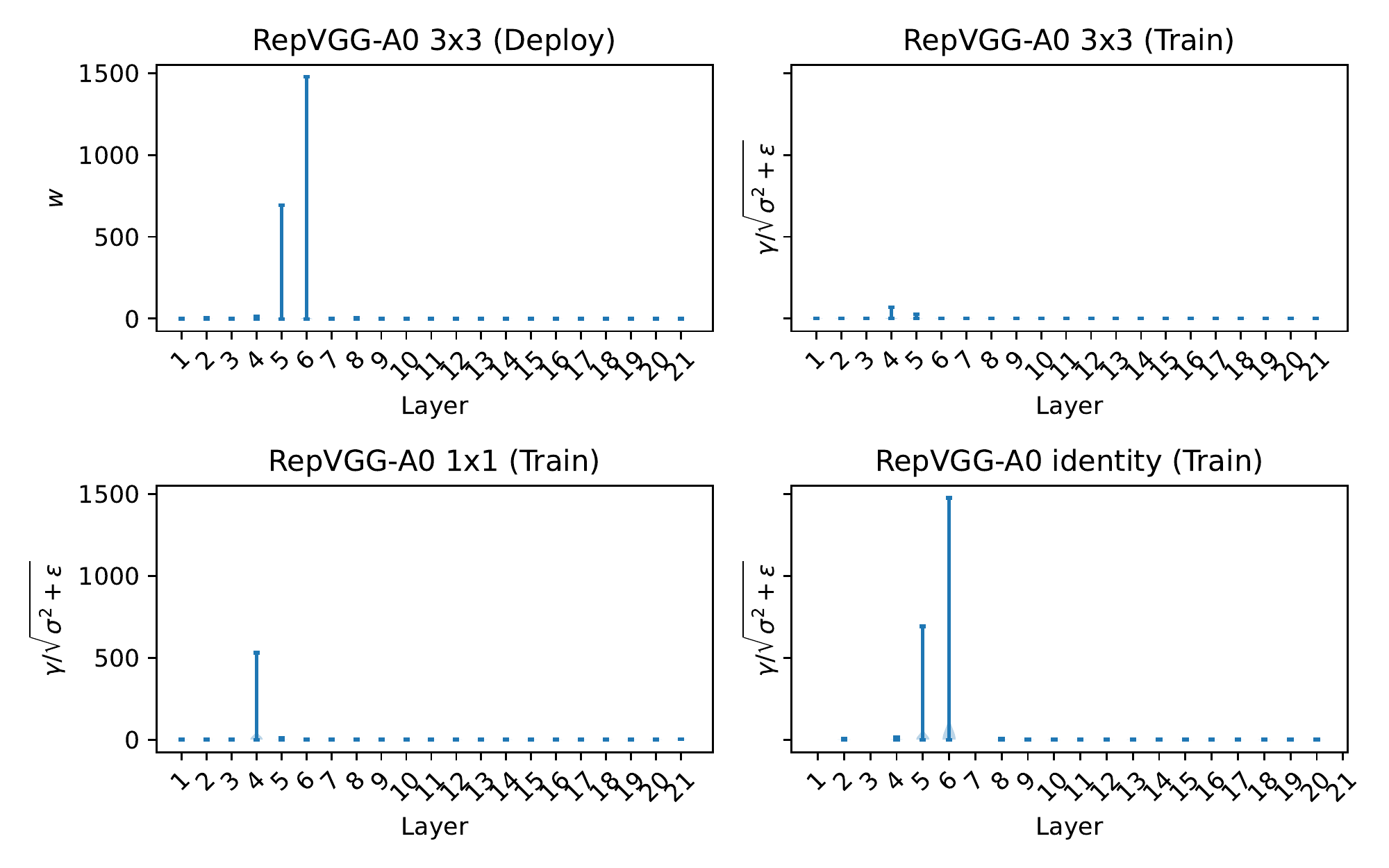}
	\caption{Violin plot of convolutional weights in each layer of RepVGG-A0  trained without custom $L_2$ (\textbf{S1}). The weight of layer 5 and 6 under the deploy setting have large variances, incurring large quantization errors (\textbf{C1 violation}).} 
	\label{fig:repvgg-a0-nocwd-w-violin}
	\vskip -0.2in
\end{figure}

\subsection{Quantization-friendly Reparameterization}


Based on  the normal $L_2$ loss, we solve the above issue by changing the reparameterization  structure. Specifically, we remove BN from the identity and 1$\times1$ branch plus appending an extra BN after the addition of  three branches. We name the network based on this basic reparameterization  structure QARepVGG. The result is shown in the bottom of Table~\ref{tab:abation_dis_den}.  As for A0 model, QARepVGG-A0 achieves 72.2$\%$ top-1 FP32  accuracy and 70.4$\%$ INT8 accuracy, which improves RepVGG by a large margin (+20.1\%).  Next, we elaborate the birth of this design and what role each component plays.

\subsubsection{Removing BN from Identity branch (M2) eliminates outlier uncontrolled weights to meet C1.} 
We name this setting \textbf{S2} and show the result in the third row of Table~\ref{tab:s1-s4}. 
The error analysis  on weight quantization (see Fig.~\ref{fig:repvgg-a0-v6-s2} in the appendix)  indicates this design indeed meets the requirements of \textbf{C1} and outlier uncontrolled weight no longer exists.  This model delivers a lower FP32 accuracy $70.7\%$ and INT8 accuracy $62.5\%$, which is still infeasible.

The error analysis  on weight quantization (see Fig.~\ref{fig:repvgg-a0-v6-s2} in the appendix)  indicates this design indeed meets the requirements of \textbf{C1}.  This model delivers a lower FP32 accuracy $70.7\%$ and INT8 accuracy $62.5\%$, which is still infeasible.  This motivates us to verify if it violates \textbf{C2}.

\subsubsection{Violating the same mean across   several branches shrinks variance of    summation to meet C2.}
\begin{figure}[ht]
	\centering
	\includegraphics[width=0.85\columnwidth]{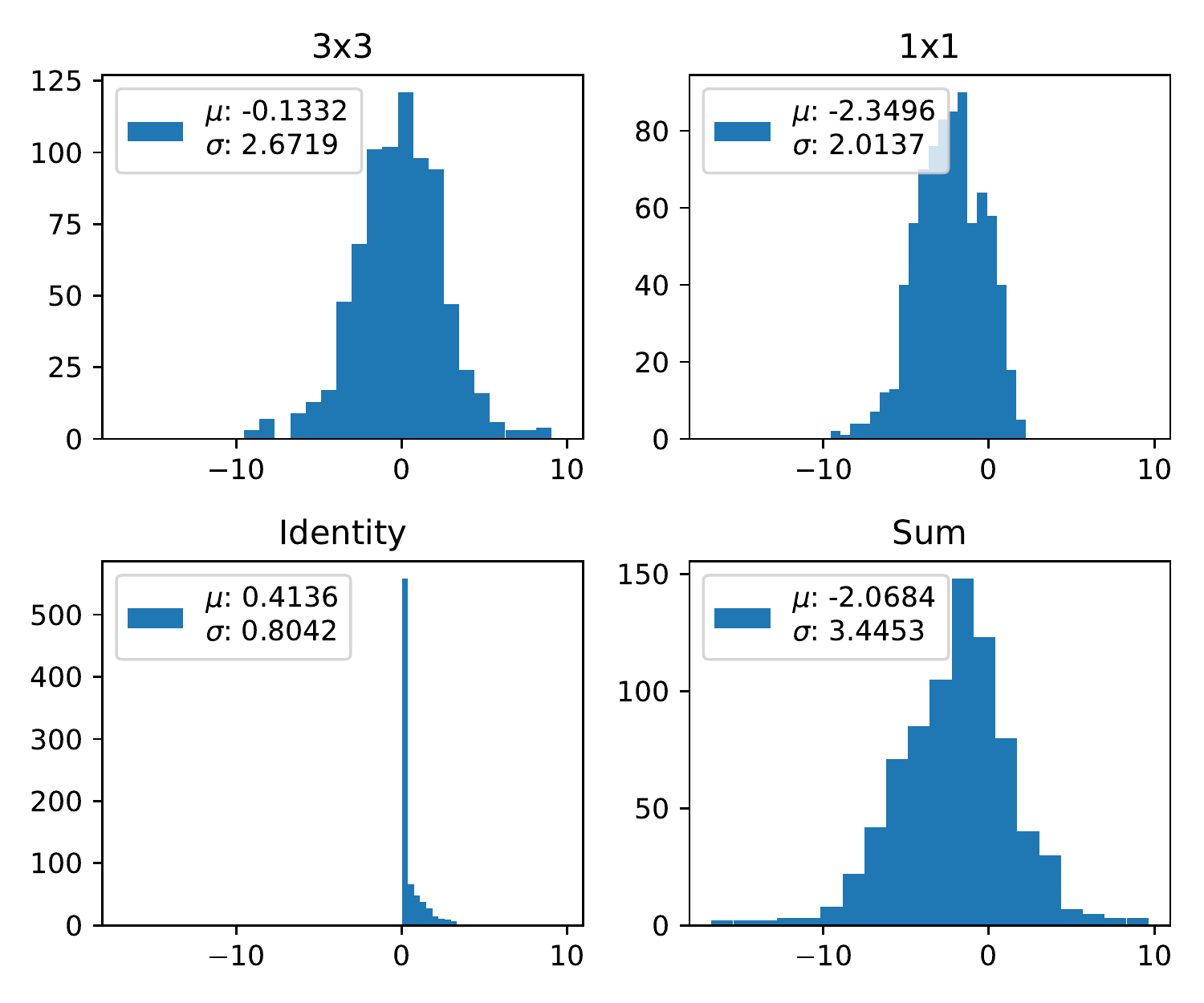}
	\caption{Enlarged variance in the activation distribution of RepVGG-A0 (\textbf{S2}). We pick a random image from the ImageNet validation set and draw the output (channel 0) of each branch at layer 4 (it is also easily seen in other layers).}
	\label{fig:qarepvgg-a0-qa-l4-c1-before-act}
	  \vskip -0.1in
\end{figure}

If the $3\times3$ and 1$\times$1 branch have the same mean, their summation is prone to enlarging the variance.  This phenomenon occurs frequently under the design of RepVGG. Specifically, ReLU \cite{nair2010rectified} is the activation function in RepVGG.
On one hand, it's harmful if most inputs are below zero (dead ReLU) \cite{maas2013rectifier}. On the other hand, it's also not favored if all inputs are above zero because of losing non-linearity. Empirically, many modern high-performance CNN models with BN often have zero means before ReLU. If we take this assumption, we would let $\mathrm{E}(\mathrm{M_{(2)}})=\mathrm{E(\vect{\mathrm{Y_{(1)}}+\mathrm{Y_{(3)}+\mathrm{Y_{(0)}}}})}=\vect{0}$.  If the $3\times3$ and 1$\times$1 branch have the same mean, we reach $\vect{\beta_{(3)}}=\vect{\beta_{(1)}}=-\frac{\mathrm{E(\vect{Y_{(0)}})}}{2}$. Note $\mathrm{E}(\vect{Y_{(0)}})\geq \vect{0}$,  adding three branches often enlarges the variance (Fig.~\ref{fig:qarepvgg-a0-qa-l4-c1-before-act}). Next, we  prove that the original design of RepVGG inevitably falls into this issue $\vect{\beta_{(3)}}=\vect{\beta_{(1)}}$ as in  Lemma~\ref{lem:m}. 
\begin{table}[h]
	\setlength{\tabcolsep}{2pt}  
	\centering
	\begin{tabular}{clcccc}
		\toprule
		Settings &  &FP32 Acc (\%)& INT8 Acc(\%) \\
		\midrule
		S0&RepVGG-A0& 72.2& 50.3\\
		S1&+\textbf{M1} &71.7& 61.6 \\
		S2&+\textbf{M1}+\textbf{M2}&70.7& 62.5\\
		S3&+\textbf{M1}+\textbf{M2}+\textbf{M3}  & 70.1& 69.5\\
		S4&+\textbf{M1}+\textbf{M2}+\textbf{M3}+\textbf{M4} & 72.2& 70.4 \\
		\midrule
		-&+\textbf{M2} &70.2 & 62.5\\
		-&+\textbf{M3} &70.4 &  69.5\\
		-&+\textbf{M4} &72.1& 57.0  \\
		\bottomrule
	\end{tabular}
	\caption{Quantization-friendly design for reparameterization and component analysis. 
	}
	\label{tab:s1-s4}
\end{table}

We write the expectation of the statistics in three branches as,

\begin{equation}
	\mathrm{E(\vect{Y_{(3)}})=\vect{\beta_{(3)}}},	\mathrm{E(\vect{Y_{(1)}})=\vect{\beta_{(1)}}}.
\end{equation}

\begin{lemma}
	\label{lem:m}
	Training a neural network using setting \textbf{S2} across $n$ iterations using loss function $l(\mathrm{W},\vect{\gamma},\vect{\beta})$ , for any given layer,   $\vect{\beta_{(3)}^n}=\vect{\beta_{(1)}^n}$. 
\end{lemma}

The proof is given in the supplementary PDF.

To better control the variance,  several simple approaches have potentials, which are shown in Table~\ref{tab:S3}. We choose this design: removing the BN in 1$\times$1 branch (\textbf{M3}) because it has the best performance.   We name this setting \textbf{S3} and show the result in Table~\ref{tab:s1-s4}. This design achieves 70.1\% top-1 FP32  and $69.5\%$ INT 8 accuracy on ImageNet, which greatly improves the quantization performance.  However, its FP32 accuracy is still low. 

\begin{table}[h]
	\setlength{\tabcolsep}{2pt}  
	\centering
	\begin{tabular}{lclc}
		\toprule
		Variants			&FP32 Acc 	& INT8 Acc	\\
		
		&   (\%) & (\%) \\
		\midrule
		1$\times$1 wo BN & 70.1& 69.5 \\
		3$\times$3 wo BN $\star$& 0.1& - & \\
		3$\times$3 wo BN,  1$\times$1 wo BN $\star$ & 0.1& -\\
		1$\times$1 with BN (affline=False)&70.1& 65.5 \\
		\bottomrule
	\end{tabular}
	\caption{Comparison using several designs using A0 model without BN in the identity branch. $\star$: stopped training because of NAN.}
	\label{tab:S3}
\end{table}
\subsubsection{Extra  BN address the  co-variate shift issue.}
\paragraph{\textbf{S4 (Post BN on S3)}}   Since the addition of  three branches introduces the covariate shift issue \cite{ioffe2015batch}, we append an extra batch normalization after the addition of three branches (\textbf{M4}) to stabilize the training process and name this setting \textbf{S4} (Fig.~\ref{fig:qarepvgg-block} right).  The  post BN doesn't affect the equivalent kernel fusion for deployment. This further boosts the FP32 accuracy of our A0 model from $70.1\%$ to $72.2\%$ on ImageNet.  Moreover, its INT8 accuracy is enhanced to $70.4\%$. 

To summarize,  combining the above four modifications together (from \textbf{M1} to \textbf{M4}) forms our QARepVGG, whose FP32 accuracy is comparable to RepVGG and INT8 performance outperforms RepVGG by a large margin.


\section{Experiment}
We mainly focus our experiments on ImageNet  dataset \cite{deng2009ImageNet}. And we verify the generalization of our method based on a recent popular detector YOLOv6 \cite{li2022yolov6}, which extensively adopts the reparameterization design and semantic segmentation. As for PTQ, we use the Pytorch-Quantization toolkit \cite{tensorrt},  which is widely used in deployment on  NVIDIA GPUs. Weights, inputs to convolution layers and full connection layers are all quantized into 8-bit, including the first and last layer. Following the default setting of Pytorch-Quantization toolkit, the quantization scheme is set to \emph{symmetric uniform}. We use the same settings and the  calibration dataset for all the quantization results, except those officially reported ones.

\subsection{ImageNet Classification}
To make fair comparisons, we strictly control the training settings as \cite{ding2021repvgg}. The detailed setting is shown in the appendix.
The results are shown in Table~\ref{tab:imaget_net_top1}. Our models achieve comparable FP32 accuracy  as RepVGG. Notably, RepVGG severely suffers from quantization, where its INT8 accuracy largely lags behind its FP32 counterpart. For example, the top-1 accuracy of RepVGG-B0 is dropped to 40.2\% from 75.1\%. In contrast,  our method exhibits strong INT8 performance, where the accuracy drops are within 2\%.

\begin{table}[ht]
	\setlength{\tabcolsep}{2pt}  
	\centering
	\begin{tabular}{lllcll}
		\toprule
		Model			&FP32  & INT8  & FPS &	Params & FLOPs \\
		& (\%) &(\%)& & (M)&(B)\\
		\midrule
		
		RepVGG-A0$^\ddagger$& 72.4 & 52.2& 3256 & 8.30 & 1.4  \\
		RepVGG-A0$^\dagger$ & 72.2& 50.3&3256 & 8.30 & 1.4\\
		QARepVGG-A0& 72.2& \textbf{70.4 }& 3256 & 8.30 & 1.4  \\
		\hline
		RepVGG-B0$^\ddagger$ & 75.1 & 40.2& 1817 & 14.33 & 3.1 \\
		QARepVGG-B0 & 74.8& \textbf{72.9}& 1817 & 14.33 & 3.1  \\
		
		\hline
		RepVGG-B1g4$^\ddagger$& 77.6 & 0.55 & 868 & 36.12 & 7.3  \\
		QARepVGG-B1g4& 77.4 &\textbf{76.5}& 868 & 36.12 & 7.3 \\
		\hline
		
		RepVGG-B1g2$^\ddagger$& 77.8&14.5& 792 & 41.36 & 8.8\\
		QARepVGG-B1g2& 77.7& \textbf{77.0}& 792 & 41.36 & 8.8  \\
		\hline 
		
		RepVGG-B1$^\ddagger$& 78.4 & 3.4& 685 & 51.82 & 11.8  \\
		QARepVGG-B1& 78.0 & \textbf{76.4}& 685 & 51.82 & 11.8  \\
		\hline
		
		RepVGG-B2g4$^\ddagger$ & 78.5 &13.7 & 581 & 55.77 & 11.3  \\
		QARepVGG-B2g4 & 78.4  &\textbf{77.7}& 581 & 55.77 & 11.3  \\
		\hline
		
		RepVGG-B2$^\ddagger$& 78.8  &51.3& 460 & 80.31 & 18.4  \\
		QARepVGG-B2& 79.0 & \textbf{77.7}&460 & 80.31 & 18.4 \\
		\bottomrule
	\end{tabular}
	\caption{Classification results on ImageNet validation dataset.  All  models are trained under the same settings and are evaluated in deploy mode.  $^\dagger$: reproduced. $^\ddagger$: RepVGG official.}
	\label{tab:imaget_net_top1}
	\vskip -0.1in
\end{table}

We observe that RepVGG with group convolutions behaves much worse. The  accuracy of RepVGG-B2g4 is dropped from 78.5\% to 13.7\%  after PTQ (64.8\%$\downarrow$). Whereas, our QARepVGG-B2g4 only loses 0.7\% accuracy, indicating its robustness to other scales and variants.

\paragraph{Comparison with RepOpt-VGG.}

RepOpt-VGG  uses gradient reparameterization and it contains two stages: searching the scales and training with the scales obtained. Quantization accuracy can be very sensitive depending on the search quality of scales \cite{ding2023reparameterizing}.

As only a few pre-trained models are released, we retrain RepOpt-VGG-A0/B0 models following \cite{ding2023reparameterizing}. Namely, we run a hyper-parameter searching for  240 epochs on CIFAR-100  and train for a complete 120 epochs on ImageNet. We can reproduce the result of RepOpt-VGG-B1 with the officially released scales. However, it was hard to find out good scales for A0/B0 to have comparable performance. As shown in Table~\ref{tab:repopt}, RepOpt-VGG-A0 achieves 70.3\% on ImageNet, which is  2.1\% lower than RepVGG. Although being much better than RepVGG, their PTQ accuracies are still too low\footnote{ See error analysis for RepOpt-VGG-B0 in Fig.~\ref{fig:repopt-vgg-b0-mse} (appendix).}.  In contrast, our method outperforms RepOpt with clear margins. Besides, we don't have sensitive hyper-parameters or extra training costs.

\begin{table}
	\setlength{\tabcolsep}{2pt}  
	\centering
	\begin{tabular}{lccc}
		\toprule
		Model			&FP32 acc & INT8 acc & Epochs	\\
		& (\%)  &(\%) & \\
		\hline
		RepOpt-VGG-A0& 70.3 & 64.8 (5.5$\downarrow$) & 240$^\ddagger$+120 \\
		QARepVGG-A0& 72.2 & 70.4 (1.8$\downarrow$) & 120\\
		\hline
		RepOpt-VGG-B0& 73.8 & 62.6 (11.2$\downarrow$) & 240$^\ddagger$+120 \\
		QARepVGG-B0& 74.8 & 72.9 (1.9$\downarrow$) & 120\\
		\hline
		RepOpt-VGG-B1$^\star$ & 78.5 & 75.9 (2.6$\downarrow$) & 240$^\ddagger$+120 \\
		RepOpt-VGG-B1$^\dagger$ & 78.3 & 75.9 (2.4$\downarrow$) & 240$^\ddagger$+120 \\
		QARepVGG-B1& 78.0 & 76.4 ($1.6\downarrow$) & 120\\
		\bottomrule
	\end{tabular}
	\caption{Comparison with RepOpt-VGG on ImageNet dataset. $^\star$: official repo.  $^\dagger$: reproduced. $^\ddagger$: 240 epochs on CIFAR-100.}
	\label{tab:repopt}
		\vskip -0.1in
\end{table}

\paragraph{Comparison using  QAT.} 

We apply QAT from the NVIDIA quantization toolkit on RepVGG, which is de facto standard in practice. The result is shown in Table~\ref{tab:pq}. While QAT  significantly boosts the quantization performance of RepVGG, it still struggles to deliver ideal performances because QAT accuracy usually matches FP32. When equipped with QAT,  QARepVGG  still outperforms RepVGG+QAT by a clear margin. As for QAT comparisons, 1\%$\uparrow$  is recognized as significant improvement.  All models are trained for 10 epochs (the first three ones for warm-up) with an initial learning rate of 0.01.

\begin{table}[h]
	\setlength{\tabcolsep}{2pt}  
	\centering
	\begin{tabular}{lccc}
		\hline
		Model			&FP32 (\%)& PTQ (\%)& QAT (\%)\\
		\hline
		RepVGG-A0 & 72.2&50.3& 66.3 \\
		QARepVGG-A0& 72.2 & 70.4  &  71.9 (\textbf{5.6$\uparrow$}) \\
		\hline
		RepVGG-B1g2 &77.8&14.5& 76.4 \\
		QARepVGG-B1g2& 77.7 & 77.0 & 77.4  (\textbf{1.0$\uparrow$})  \\
		\hline
		RepVGG-B2 &78.8&51.3& 77.4 \\
		QARepVGG-B2& 79.0 & 77.7 &  78.7 (\textbf{1.3$\uparrow$})  \\
		\hline
	\end{tabular}
	\vskip -0.1in
	\caption{PTQ and QAT results on ImageNet validation set.}
	\label{tab:pq}
\end{table}


\paragraph{Ablation study on component analysis.}
We study the contribution to  quantization performance from four modifications and show the results in Table~\ref{tab:s1-s4}. Note that when BN is entirely removed, the model fails to converge. Our design, putting these four components together, is deduced by meeting both \textbf{C1} and \textbf{C2} requirements. Single component analysis helps to evaluate which role it plays more quantitatively.  It's interesting that the  \textbf{M3} setting is very near to the VGG-BN-A0 setting (the second row of Table~\ref{tab:imaget_net_top1_vgg}), which has  lower FP32 and relative higher INT8 accuracy. However, our fully equipped QARepVGG achieves the best FP32 and INT8 accuracy simultaneously.
\subsection{Object Detection}
To further verify the generalization of QARepVGG, we test it on object detectors like YOLOv6 \cite{li2022yolov6}. It extensively makes use of RepVGG blocks and severely suffers from the quantization issue. Although YOLOv6 alleviates this issue by resorting to RepOpt-VGG, the approach is unstable and requires very careful hyperparameter tuning.  

We take `tiny' and `small' model variants as comparison benchmarks. We train and evaluate QARepVGG-fashioned YOLOv6    on the COCO 2017 dataset \cite{lin2014microsoft} and exactly follow its official settings \cite{li2022yolov6}. The results are shown in Table~\ref{tab:yolov6}. RepVGG and QARepVGG versions are trained for 300 epochs on 8 Tesla-V100 GPUs. RepOpt requires extra 300 epochs to search for scales. 

Noticeably, YOLOv6s-RepVGG suffers a huge quantization degradation for about 7.4\% mAP via PTQ. YOLOv6t-RepVGG is slightly better, but the reduction of 3\% mAP is again unacceptable in practical deployment. Contrarily, YOLOv6s/t-QARepVGG have similar FP32 accuracies to their RepVGG counterpart, while INT8 accuracy drops are restricted within 1.3\% mAP. YOLOv6-RepOpt-VGG could give better PTQ accuracy than YOLOv6-RepVGG as well. However, it requires a doubled cost. We also find that the final accuracy of RepOpt-VGG is quite sensitive to the searched hyper-parameters which cannot be robustly obtained.
\begin{table}
	\setlength{\tabcolsep}{2pt}  
	\centering
	\begin{tabular}{lccc}
		\toprule
		Model			&FP32 mAP & INT8 mAP & Epochs	\\
		& (\%)  &(\%) &  \\
		\hline
		
		YOLOv6t-RepVGG& 40.8 & 37.8 (3.0$\downarrow$) & 300 \\
		YOLOv6t-RepOpt& 40.7 & 39.1 (1.6$\downarrow$) & 300+300\\
		YOLOv6t-QARepVGG& 40.7 & \textbf{39.5 (1.2$\downarrow$)}& 300\\
		\hline
		YOLOv6s-RepVGG$^\star$ & 42.4& 35.0 (7.4$\downarrow$)&300\\
		YOLOv6s-RepOpt$^\star$&  42.4& 40.9 (1.5$\downarrow$)& 300+300\\
		YOLOv6s-QARepVGG& 42.3 & \textbf{41.0 (1.3$\downarrow$)}& 300\\
		\bottomrule
	\end{tabular}
	\caption{Detection results on COCO.  $^\star$:  official repo.}
	\label{tab:yolov6}
\end{table}



\subsection{Semantic Segmentation}

We further evaluate our method on the semantic segmentation task. Specifically, we use two representative frameworks FCN and DeepLabV3+ \cite{deeplabv3plus2018} The detailed setting is shown in the supplementary.  The results are shown in Table~\ref{tab:seg}. Under the FCN framework, the mIoU is reduced from 72.5\% (fp32) to 67.1\% (int8)  using RepVGG-B1g4. In contrast, mIoU is reduced from 72.6\% (fp32)  to 71.4\% (int8) on top of QARepVGG-B1g4. Under the DeepLabv3+ framework, RepVGG-B1g4 severely suffers from the quantization with 5.3\% mIoU drop. Whereas, QARepVGG-B1g4 only drops 1.2\%.

\begin{table}
	\setlength{\tabcolsep}{2pt}  
	\centering
		%
	\begin{tabular}{lcc}
		\toprule
		Model		 	&mIoU & mIoU\\
		&  FP32(\%)  &INT8(\%)   \\
		\hline
		FCN(RepVGG-B1g4)&   72.5 & 67.1\\
		FCN(QARepVGG-B1g4)& 72.6& \textbf{71.4} \\
		
		\hline
		DeepLabV3+(RepVGG-B1g4) & 78.4& 73.1\\
		DeepLabV3+(QARepVGG-B1g4) &78.4& \textbf{77.2} \\
		\bottomrule
	\end{tabular}
	\caption{Semantic segmentation results on cityscapes. All models are trained using crop size of 512$\times$1024.
	}
	\label{tab:seg}
		\vskip -0.15in
\end{table}

\section{Conclusion}
Through theoretical and quantitative analysis, we dissect the well-known quantization failure of the notable reparameterization-based structure RepVGG. Its structural defect inevitably magnifies the quantization error and cumulatively produces inferior results. We refashion its design to have QARepVGG, which generates the weight and activation distributions that are advantageous for quantization. While keeping the good FP32 performance of RepVGG,  QARepVGG greatly eases the quantization process for final deployment. We emphasize that quantization awareness in architectural design shall be drawn more attention. 

\textbf{Acknowledgements}: This work was supported by National Key R\&D Program of China (No. 2022ZD0118700).
\bibliography{egbib}

\begin{thebibliography}{51}
\providecommand{\natexlab}[1]{#1}

\bibitem[{Bhalgat et~al.(2020)Bhalgat, Lee, Nagel, Blankevoort, and
  Kwak}]{bhalgat2020lsq+}
Bhalgat, Y.; Lee, J.; Nagel, M.; Blankevoort, T.; and Kwak, N. 2020.
\newblock Lsq+: Improving low-bit quantization through learnable offsets and
  better initialization.
\newblock In \emph{Proceedings of the IEEE/CVF Conference on Computer Vision
  and Pattern Recognition Workshops}, 696--697.

\bibitem[{Cai et~al.(2020)Cai, Yao, Dong, Gholami, Mahoney, and
  Keutzer}]{cai2020zeroq}
Cai, Y.; Yao, Z.; Dong, Z.; Gholami, A.; Mahoney, M.~W.; and Keutzer, K. 2020.
\newblock Zeroq: A novel zero shot quantization framework.
\newblock In \emph{Proceedings of the IEEE/CVF Conference on Computer Vision
  and Pattern Recognition}, 13169--13178.

\bibitem[{Chen et~al.(2017)Chen, Papandreou, Kokkinos, Murphy, and
  Yuille}]{chen2017deeplab}
Chen, L.-C.; Papandreou, G.; Kokkinos, I.; Murphy, K.; and Yuille, A.~L. 2017.
\newblock Deeplab: Semantic image segmentation with deep convolutional nets,
  atrous convolution, and fully connected crfs.
\newblock \emph{IEEE transactions on pattern analysis and machine
  intelligence}, 40(4): 834--848.

\bibitem[{Chen et~al.(2018)Chen, Zhu, Papandreou, Schroff, and
  Adam}]{deeplabv3plus2018}
Chen, L.-C.; Zhu, Y.; Papandreou, G.; Schroff, F.; and Adam, H. 2018.
\newblock Encoder-Decoder with Atrous Separable Convolution for Semantic Image
  Segmentation.
\newblock In \emph{ECCV}.

\bibitem[{Cordts et~al.(2016)Cordts, Omran, Ramos, Rehfeld, Enzweiler,
  Benenson, Franke, Roth, and Schiele}]{cordts2016cityscapes}
Cordts, M.; Omran, M.; Ramos, S.; Rehfeld, T.; Enzweiler, M.; Benenson, R.;
  Franke, U.; Roth, S.; and Schiele, B. 2016.
\newblock The cityscapes dataset for semantic urban scene understanding.
\newblock In \emph{Proceedings of the IEEE conference on computer vision and
  pattern recognition}, 3213--3223.

\bibitem[{Dehner et~al.(2016)Dehner, Dehner, Rabenstein, Sch{\"a}fer, and
  Strobl}]{dehner2016analysis}
Dehner, G.; Dehner, I.; Rabenstein, R.; Sch{\"a}fer, M.; and Strobl, C. 2016.
\newblock Analysis of the quantization error in digital multipliers with small
  wordlength.
\newblock In \emph{2016 24th European Signal Processing Conference (EUSIPCO)},
  1848--1852. IEEE.

\bibitem[{Deng et~al.(2009)Deng, Dong, Socher, Li, Li, and
  Fei-Fei}]{deng2009ImageNet}
Deng, J.; Dong, W.; Socher, R.; Li, L.-J.; Li, K.; and Fei-Fei, L. 2009.
\newblock Imagenet: A large-scale hierarchical image database.
\newblock In \emph{2009 IEEE conference on computer vision and pattern
  recognition}, 248--255. Ieee.

\bibitem[{Devlin et~al.(2019)Devlin, Chang, Lee, and
  Toutanova}]{devlin2019bert}
Devlin, J.; Chang, M.-W.; Lee, K.; and Toutanova, K. 2019.
\newblock BERT: Pre-training of Deep Bidirectional Transformers for Language
  Understanding.
\newblock In \emph{Proceedings of the 2019 Conference of the North American
  Chapter of the Association for Computational Linguistics: Human Language
  Technologies, Volume 1 (Long and Short Papers)}, 4171--4186.

\bibitem[{Ding et~al.(2023)Ding, Chen, Zhang, Huang, Han, and
  Ding}]{ding2023reparameterizing}
Ding, X.; Chen, H.; Zhang, X.; Huang, K.; Han, J.; and Ding, G. 2023.
\newblock Re-parameterizing Your Optimizers rather than Architectures.
\newblock In \emph{The Eleventh International Conference on Learning
  Representations}.

\bibitem[{Ding et~al.(2019)Ding, Guo, Ding, and Han}]{ding2019acnet}
Ding, X.; Guo, Y.; Ding, G.; and Han, J. 2019.
\newblock Acnet: Strengthening the kernel skeletons for powerful cnn via
  asymmetric convolution blocks.
\newblock In \emph{Proceedings of the IEEE/CVF international conference on
  computer vision}, 1911--1920.

\bibitem[{Ding et~al.(2022)Ding, Zhang, Han, and Ding}]{ding2022scaling}
Ding, X.; Zhang, X.; Han, J.; and Ding, G. 2022.
\newblock Scaling up your kernels to 31x31: Revisiting large kernel design in
  cnns.
\newblock In \emph{Proceedings of the IEEE/CVF Conference on Computer Vision
  and Pattern Recognition}, 11963--11975.

\bibitem[{Ding et~al.(2021)Ding, Zhang, Ma, Han, Ding, and
  Sun}]{ding2021repvgg}
Ding, X.; Zhang, X.; Ma, N.; Han, J.; Ding, G.; and Sun, J. 2021.
\newblock Repvgg: Making vgg-style convnets great again.
\newblock In \emph{Proceedings of the IEEE/CVF Conference on Computer Vision
  and Pattern Recognition}, 13733--13742.
\newblock \url{https://github.com/DingXiaoH/RepVGG.git}, hashtag:
  5c2e359a144726b9d14cba1e455bf540eaa54afc.

\bibitem[{Dosovitskiy et~al.(2020)Dosovitskiy, Beyer, Kolesnikov, Weissenborn,
  Zhai, Unterthiner, Dehghani, Minderer, Heigold, Gelly
  et~al.}]{dosovitskiy2020image}
Dosovitskiy, A.; Beyer, L.; Kolesnikov, A.; Weissenborn, D.; Zhai, X.;
  Unterthiner, T.; Dehghani, M.; Minderer, M.; Heigold, G.; Gelly, S.; et~al.
  2020.
\newblock An Image is Worth 16x16 Words: Transformers for Image Recognition at
  Scale.
\newblock In \emph{International Conference on Learning Representations}.

\bibitem[{Graves, Mohamed, and Hinton(2013)}]{graves2013speech}
Graves, A.; Mohamed, A.-r.; and Hinton, G. 2013.
\newblock Speech recognition with deep recurrent neural networks.
\newblock In \emph{2013 IEEE international conference on acoustics, speech and
  signal processing}, 6645--6649. Ieee.

\bibitem[{Gupta et~al.(2015)Gupta, Agrawal, Gopalakrishnan, and
  Narayanan}]{gupta2015deep}
Gupta, S.; Agrawal, A.; Gopalakrishnan, K.; and Narayanan, P. 2015.
\newblock Deep learning with limited numerical precision.
\newblock In \emph{International conference on machine learning}, 1737--1746.
  PMLR.

\bibitem[{Gysel et~al.(2018)Gysel, Pimentel, Motamedi, and
  Ghiasi}]{gysel2018ristretto}
Gysel, P.; Pimentel, J.; Motamedi, M.; and Ghiasi, S. 2018.
\newblock Ristretto: A framework for empirical study of resource-efficient
  inference in convolutional neural networks.
\newblock \emph{IEEE transactions on neural networks and learning systems},
  29(11): 5784--5789.

\bibitem[{Habi et~al.(2021)Habi, Peretz, Cohen, Dikstein, Dror, Diamant,
  Jennings, and Netzer}]{habi2021hptq}
Habi, H.~V.; Peretz, R.; Cohen, E.; Dikstein, L.; Dror, O.; Diamant, I.;
  Jennings, R.~H.; and Netzer, A. 2021.
\newblock HPTQ: Hardware-Friendly Post Training Quantization.
\newblock \emph{arXiv preprint arXiv:2109.09113}.

\bibitem[{He et~al.(2017)He, Gkioxari, Doll{\'a}r, and Girshick}]{he2017mask}
He, K.; Gkioxari, G.; Doll{\'a}r, P.; and Girshick, R. 2017.
\newblock Mask r-cnn.
\newblock In \emph{Proceedings of the IEEE international conference on computer
  vision}, 2961--2969.

\bibitem[{He et~al.(2016)He, Zhang, Ren, and Sun}]{he2016deep}
He, K.; Zhang, X.; Ren, S.; and Sun, J. 2016.
\newblock Deep residual learning for image recognition.
\newblock In \emph{Proceedings of the IEEE conference on computer vision and
  pattern recognition}, 770--778.

\bibitem[{Howard et~al.(2019)Howard, Sandler, Chu, Chen, Chen, Tan, Wang, Zhu,
  Pang, Vasudevan et~al.}]{howard2019searching}
Howard, A.; Sandler, M.; Chu, G.; Chen, L.-C.; Chen, B.; Tan, M.; Wang, W.;
  Zhu, Y.; Pang, R.; Vasudevan, V.; et~al. 2019.
\newblock Searching for mobilenetv3.
\newblock In \emph{Proceedings of the IEEE/CVF international conference on
  computer vision}, 1314--1324.

\bibitem[{Howard et~al.(2017)Howard, Zhu, Chen, Kalenichenko, Wang, Weyand,
  Andreetto, and Adam}]{howard2017mobilenets}
Howard, A.~G.; Zhu, M.; Chen, B.; Kalenichenko, D.; Wang, W.; Weyand, T.;
  Andreetto, M.; and Adam, H. 2017.
\newblock Mobilenets: Efficient convolutional neural networks for mobile vision
  applications.
\newblock \emph{arXiv preprint arXiv:1704.04861}.

\bibitem[{Hu et~al.(2022)Hu, Feng, Hua, Lai, Huang, Gong, and
  Hua}]{hu2022online}
Hu, M.; Feng, J.; Hua, J.; Lai, B.; Huang, J.; Gong, X.; and Hua, X.-S. 2022.
\newblock Online Convolutional Re-parameterization.
\newblock In \emph{Proceedings of the IEEE/CVF Conference on Computer Vision
  and Pattern Recognition}, 568--577.

\bibitem[{Huang et~al.(2022{\natexlab{a}})Huang, You, Zhang, Du, Wang, Qian,
  and Xu}]{huang2022dyrep}
Huang, T.; You, S.; Zhang, B.; Du, Y.; Wang, F.; Qian, C.; and Xu, C.
  2022{\natexlab{a}}.
\newblock DyRep: Bootstrapping Training with Dynamic Re-parameterization.
\newblock In \emph{Proceedings of the IEEE/CVF Conference on Computer Vision
  and Pattern Recognition}, 588--597.

\bibitem[{Huang et~al.(2022{\natexlab{b}})Huang, Zhang, Heng, Shi, and
  Zhou}]{huang2022rife}
Huang, Z.; Zhang, T.; Heng, W.; Shi, B.; and Zhou, S. 2022{\natexlab{b}}.
\newblock Real-Time Intermediate Flow Estimation for Video Frame Interpolation.
\newblock In \emph{Proceedings of the European Conference on Computer Vision
  (ECCV)}.

\bibitem[{Ignatov et~al.(2023)Ignatov, Timofte, Denna, Younes, Gankhuyag, Huh,
  Kim, Yoon, Moon, Lee et~al.}]{ignatov2023efficient}
Ignatov, A.; Timofte, R.; Denna, M.; Younes, A.; Gankhuyag, G.; Huh, J.; Kim,
  M.~K.; Yoon, K.; Moon, H.-C.; Lee, S.; et~al. 2023.
\newblock Efficient and accurate quantized image super-resolution on mobile
  NPUs, mobile AI \& AIM 2022 challenge: report.
\newblock In \emph{Computer Vision--ECCV 2022 Workshops: Tel Aviv, Israel,
  October 23--27, 2022, Proceedings, Part III}, 92--129. Springer.

\bibitem[{Ioffe and Szegedy(2015)}]{ioffe2015batch}
Ioffe, S.; and Szegedy, C. 2015.
\newblock Batch normalization: Accelerating deep network training by reducing
  internal covariate shift.
\newblock In \emph{International conference on machine learning}, 448--456.
  PMLR.

\bibitem[{Jacob et~al.(2018)Jacob, Kligys, Chen, Zhu, Tang, Howard, Adam, and
  Kalenichenko}]{jacob2018quantization}
Jacob, B.; Kligys, S.; Chen, B.; Zhu, M.; Tang, M.; Howard, A.; Adam, H.; and
  Kalenichenko, D. 2018.
\newblock Quantization and training of neural networks for efficient
  integer-arithmetic-only inference.
\newblock In \emph{Proceedings of the IEEE conference on computer vision and
  pattern recognition}, 2704--2713.

\bibitem[{Krishnamoorthi(2018)}]{krishnamoorthi2018quantizing}
Krishnamoorthi, R. 2018.
\newblock Quantizing deep convolutional networks for efficient inference: A
  whitepaper.
\newblock \emph{arXiv preprint arXiv:1806.08342}.

\bibitem[{Krizhevsky, Hinton et~al.(2009)}]{krizhevsky2009learning}
Krizhevsky, A.; Hinton, G.; et~al. 2009.
\newblock Learning multiple layers of features from tiny images.

\bibitem[{Li et~al.(2022)Li, Li, Jiang, Weng, Geng, Li, Ke, Li, Cheng, Nie, Li,
  Zhang, Liang, Zhou, Xu, Chu, Wei, and Wei}]{li2022yolov6}
Li, C.; Li, L.; Jiang, H.; Weng, K.; Geng, Y.; Li, L.; Ke, Z.; Li, Q.; Cheng,
  M.; Nie, W.; Li, Y.; Zhang, B.; Liang, Y.; Zhou, L.; Xu, X.; Chu, X.; Wei,
  X.; and Wei, X. 2022.
\newblock YOLOv6: a single-stage object detection framework for industrial
  applications.
\newblock \emph{arXiv preprint arXiv:2209.02976}.
\newblock \url{https://github.com/meituan/YOLOv6.git}, hashtag:
  05da1477671017ac2edbb709e09c75854a7b4eb1.

\bibitem[{Lin et~al.(2014)Lin, Maire, Belongie, Hays, Perona, Ramanan,
  Doll{\'a}r, and Zitnick}]{lin2014microsoft}
Lin, T.-Y.; Maire, M.; Belongie, S.; Hays, J.; Perona, P.; Ramanan, D.;
  Doll{\'a}r, P.; and Zitnick, C.~L. 2014.
\newblock Microsoft coco: Common objects in context.
\newblock In \emph{European conference on computer vision}, 740--755. Springer.

\bibitem[{Long, Shelhamer, and Darrell(2015)}]{long2015fully}
Long, J.; Shelhamer, E.; and Darrell, T. 2015.
\newblock Fully convolutional networks for semantic segmentation.
\newblock In \emph{Proceedings of the IEEE conference on computer vision and
  pattern recognition}, 3431--3440.

\bibitem[{Maas et~al.(2013)Maas, Hannun, Ng et~al.}]{maas2013rectifier}
Maas, A.~L.; Hannun, A.~Y.; Ng, A.~Y.; et~al. 2013.
\newblock Rectifier nonlinearities improve neural network acoustic models.
\newblock In \emph{ICML}, volume~30, 3. Atlanta, Georgia, USA.

\bibitem[{Nagel et~al.(2019)Nagel, Baalen, Blankevoort, and
  Welling}]{nagel2019data}
Nagel, M.; Baalen, M.~v.; Blankevoort, T.; and Welling, M. 2019.
\newblock Data-free quantization through weight equalization and bias
  correction.
\newblock In \emph{Proceedings of the IEEE/CVF International Conference on
  Computer Vision}, 1325--1334.

\bibitem[{Nair and Hinton(2010)}]{nair2010rectified}
Nair, V.; and Hinton, G.~E. 2010.
\newblock Rectified linear units improve restricted boltzmann machines.
\newblock In \emph{ICML}.

\bibitem[{NVIDIA(2018)}]{tensorrt}
NVIDIA. 2018.
\newblock {TensorRT}.
\newblock \url{https://developer.nvidia.com/tensorrt}.

\bibitem[{Redmon et~al.(2016)Redmon, Divvala, Girshick, and
  Farhadi}]{redmon2016you}
Redmon, J.; Divvala, S.; Girshick, R.; and Farhadi, A. 2016.
\newblock You only look once: Unified, real-time object detection.
\newblock In \emph{Proceedings of the IEEE conference on computer vision and
  pattern recognition}, 779--788.

\bibitem[{Sandler et~al.(2018)Sandler, Howard, Zhu, Zhmoginov, and
  Chen}]{sandler2018mobilenetv2}
Sandler, M.; Howard, A.; Zhu, M.; Zhmoginov, A.; and Chen, L.-C. 2018.
\newblock Mobilenetv2: Inverted residuals and linear bottlenecks.
\newblock In \emph{Proceedings of the IEEE conference on computer vision and
  pattern recognition}, 4510--4520.

\bibitem[{Sheng et~al.(2018)Sheng, Feng, Zhuo, Zhang, Shen, and
  Aleksic}]{sheng2018quantization}
Sheng, T.; Feng, C.; Zhuo, S.; Zhang, X.; Shen, L.; and Aleksic, M. 2018.
\newblock A quantization-friendly separable convolution for mobilenets.
\newblock In \emph{2018 1st Workshop on Energy Efficient Machine Learning and
  Cognitive Computing for Embedded Applications (EMC2)}, 14--18. IEEE.

\bibitem[{Tan and Le(2019)}]{tan2019efficientnet}
Tan, M.; and Le, Q. 2019.
\newblock Efficientnet: Rethinking model scaling for convolutional neural
  networks.
\newblock In \emph{International Conference on Machine Learning}, 6105--6114.
  PMLR.

\bibitem[{Vasu et~al.(2022)Vasu, Gabriel, Zhu, Tuzel, and
  Ranjan}]{vasu2022improved}
Vasu, P. K.~A.; Gabriel, J.; Zhu, J.; Tuzel, O.; and Ranjan, A. 2022.
\newblock An Improved One millisecond Mobile Backbone.
\newblock \emph{arXiv preprint arXiv:2206.04040}.

\bibitem[{Vaswani et~al.(2017)Vaswani, Shazeer, Parmar, Uszkoreit, Jones,
  Gomez, Kaiser, and Polosukhin}]{vaswani2017attention}
Vaswani, A.; Shazeer, N.; Parmar, N.; Uszkoreit, J.; Jones, L.; Gomez, A.~N.;
  Kaiser, {\L}.; and Polosukhin, I. 2017.
\newblock Attention is all you need.
\newblock \emph{Advances in neural information processing systems}, 30.

\bibitem[{Wang, Bochkovskiy, and Liao(2022)}]{wang2022yolov7}
Wang, C.-Y.; Bochkovskiy, A.; and Liao, H.-Y.~M. 2022.
\newblock YOLOv7: Trainable bag-of-freebies sets new state-of-the-art for
  real-time object detectors.
\newblock \emph{arXiv preprint arXiv:2207.02696}.

\bibitem[{Wang et~al.(2019)Wang, Liu, Lin, Lin, and Han}]{wang2019haq}
Wang, K.; Liu, Z.; Lin, Y.; Lin, J.; and Han, S. 2019.
\newblock Haq: Hardware-aware automated quantization with mixed precision.
\newblock In \emph{Proceedings of the IEEE/CVF Conference on Computer Vision
  and Pattern Recognition}, 8612--8620.

\bibitem[{Wu et~al.(2018)Wu, Wang, Zhang, Tian, Vajda, and
  Keutzer}]{wu2018mixed}
Wu, B.; Wang, Y.; Zhang, P.; Tian, Y.; Vajda, P.; and Keutzer, K. 2018.
\newblock Mixed precision quantization of convnets via differentiable neural
  architecture search.
\newblock \emph{arXiv preprint arXiv:1812.00090}.

\bibitem[{Wu, Lee, and Ma(2022)}]{pmlr-v162-wu22f}
Wu, K.; Lee, C.-K.; and Ma, K. 2022.
\newblock {M}em{SR}: Training Memory-efficient Lightweight Model for Image
  Super-Resolution.
\newblock In Chaudhuri, K.; Jegelka, S.; Song, L.; Szepesvari, C.; Niu, G.; and
  Sabato, S., eds., \emph{Proceedings of the 39th International Conference on
  Machine Learning}, volume 162 of \emph{Proceedings of Machine Learning
  Research}, 24076--24092. PMLR.

\bibitem[{Xu et~al.(2022)Xu, Wang, Lv, Chang, Cui, Deng, Wang, Dang, Wei, Du
  et~al.}]{xu2022pp}
Xu, S.; Wang, X.; Lv, W.; Chang, Q.; Cui, C.; Deng, K.; Wang, G.; Dang, Q.;
  Wei, S.; Du, Y.; et~al. 2022.
\newblock PP-YOLOE: An evolved version of YOLO.
\newblock \emph{arXiv preprint arXiv:2203.16250}.

\bibitem[{Yun and Wong(2021)}]{yun2021all}
Yun, S.; and Wong, A. 2021.
\newblock Do All MobileNets Quantize Poorly? Gaining Insights into the Effect
  of Quantization on Depthwise Separable Convolutional Networks Through the
  Eyes of Multi-scale Distributional Dynamics.
\newblock In \emph{Proceedings of the IEEE/CVF Conference on Computer Vision
  and Pattern Recognition}, 2447--2456.

\bibitem[{Zagoruyko and Komodakis(2017)}]{zagoruyko2017diracnets}
Zagoruyko, S.; and Komodakis, N. 2017.
\newblock Diracnets: Training very deep neural networks without
  skip-connections.
\newblock \emph{arXiv preprint arXiv:1706.00388}.

\bibitem[{Zhang, Zeng, and Zhang(2021)}]{zhang2021edge}
Zhang, X.; Zeng, H.; and Zhang, L. 2021.
\newblock Edge-oriented Convolution Block for Real-time Super Resolution on
  Mobile Devices.
\newblock In \emph{Proceedings of the 29th ACM International Conference on
  Multimedia}, 4034--4043.

\bibitem[{Zhou et~al.(2023)Zhou, Tian, Chu, Zhang, Zhang, Lu, Feng, Jie,
  Chiang, and Ma}]{zhou2023fastpillars}
Zhou, S.; Tian, Z.; Chu, X.; Zhang, X.; Zhang, B.; Lu, X.; Feng, C.; Jie, Z.;
  Chiang, P.~Y.; and Ma, L. 2023.
\newblock FastPillars: A Deployment-friendly Pillar-based 3D Detector.
\newblock \emph{arXiv preprint arXiv:2302.02367}.

\end{thebibliography}

\appendix
\section{Algorithm}

We give the custom $L_2$ implementation in Alg.~\ref{alg:customL2}.
\begin{algorithm}
	\caption{The code of custom $L_2$ in RepVGG block.}
	\label{alg:customL2}
	\definecolor{codeblue}{rgb}{0.25,0.5,0.5}
	\lstset{
		backgroundcolor=\color{white},
		basicstyle=\fontsize{7.2pt}{7.2pt}\ttfamily\selectfont,
		columns=fullflexible,
		breaklines=true,
		captionpos=b,
		commentstyle=\fontsize{7.2pt}{7.2pt}\color{codeblue},
		keywordstyle=\fontsize{7.2pt}{7.2pt}\color{blue},
	}
	\begin{lstlisting}[language=python]
		def get_custom_L2(self):
		K3 = self.rbr_dense.conv.weight # 3x3
		K1 = self.rbr_1x1.conv.weight # 1x1
		t3 = (self.rbr_dense.bn.weight / ((self.rbr_dense.bn.running_var + self.rbr_dense.bn.eps).sqrt())).reshape(-1, 1, 1, 1).detach()
		t1 = (self.rbr_1x1.bn.weight / ((self.rbr_1x1.bn.running_var + self.rbr_1x1.bn.eps).sqrt())).reshape(-1, 1, 1, 1).detach()
		l2_loss_circle = (K3 ** 2).sum() - (K3[:, :, 1:2, 1:2] ** 2).sum()    
		eq_kernel = K3[:, :, 1:2, 1:2] * t3 + K1 * t1                  
		l2_loss_eq_kernel = (eq_kernel ** 2 / (t3 ** 2 + t1 ** 2)).sum()    # Normalize
		return l2_loss_eq_kernel + l2_loss_circle
		
	\end{lstlisting}
\end{algorithm}

\section{More Related Work}
\textbf{Quantization-aware architecture design.} 
A quantization-friendly replacement of separable convolution is proposed in \cite{sheng2018quantization}, where a metric called  signal-to-quantization-noise ratio (SQNR) is defined to diagnose the quantization loss of each component of the network. It is also argued that weights shall obey a uniform distribution to facilitate quantization \cite{sheng2018quantization}. Swish-like activations are known to have quantization collapse which either requires a delicate learnable quantization scheme to recover \cite{bhalgat2020lsq+}, or to be replaced by RELU6 as in EfficientNet-Lite \cite{liu2020higher}. BatchQuant \cite{bai2021batchquant} utilizes one-shot neural architecture search for robust mixed-precision models without retraining.

\section{Training Setting}
\label{sec:setting_imagenet}

\paragraph{ImageNet classfication.}As for model specification, we use the same configuration as \cite{ding2021repvgg} except for the quantization-friendly reparameterization design, which includes the A and B series. This naturally generates the same deploy model for inference.  All models are trained for 120 epochs with a global batch size of 256.  We use SGD optimizer with a momentum of 0.9 and a weight decay of $10^{-4}$. The learning rate is initialized as 0.1 and decayed to zero following a cosine strategy.  We also follow the simple data augmentations as \cite{ding2021repvgg}. All experiments are done on 8 Tesla-V100 GPUs. 

\paragraph{Semantic segmentation.}  We utilize the  mmsegmentation  \cite{mmseg2020} framework. The hyper-parameters follows the default setting in mmsegmentation. We use SGD with 0.9 momentum and a global batch size of 16. The initial learning rate is 0.1 and decayed to 0.0001 following the polynomial strategy. We use 40k and 80k iterations for FCN and DeepLabv3+ respectively.

\section{Discussion} 
Our method is a structural cure to RepVGG's quantization difficulty that both works for PTQ  and QAT.  Though QAT generally delivers better performance, we primarily focus on PTQ since it is more widely used and there is no extra training cost, for this reason we recommended it as the default quantization method. 
We emphasize that \emph{QAT methods can only be applied for RepVGG in deploy mode} (single-branch), otherwise directly quantizing multi-branch modules makes them difficult to be fused for higher throughput,  where each branch usually has different input ranges and activation distributions. This fact largely limits the feasible QAT performance of RepVGG. 

\section{More experiments}

\paragraph{Error Analysis.}

\begin{figure}[ht]
	\centering
	\includegraphics[width=0.9\columnwidth]{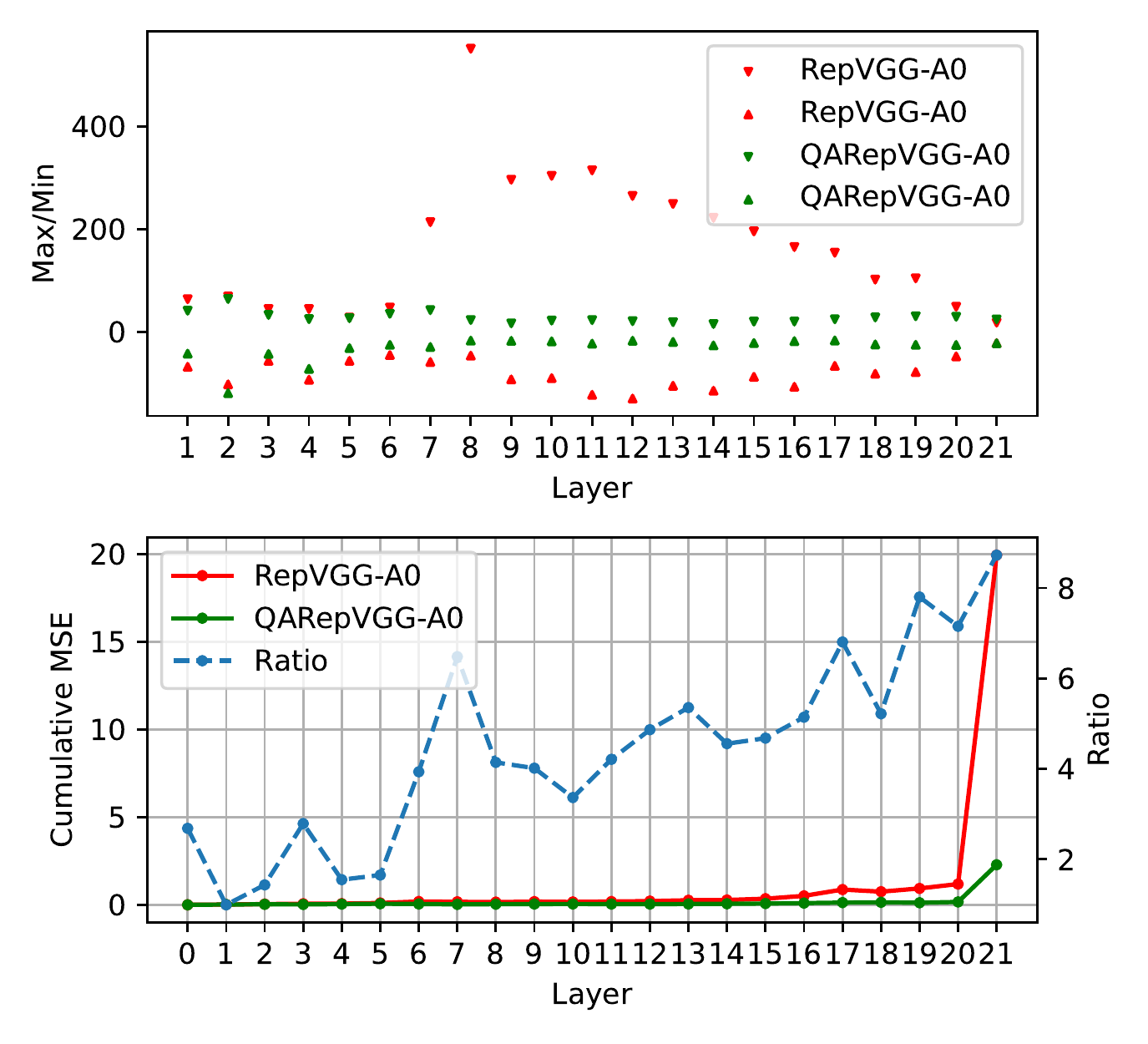}
	\caption{\textbf{Top}: max/min value of feature maps (the same batch is used) before activation of each QARepVGG-A0 block  compared with RepVGG-A0. \textbf{Bottom}: Cumulative MSE is tested by sequentially switching on layers one after another in INT8. The blue-dashed line indicates the relative ratio between the two.}
	\label{fig:qarepvgg-a0-before-act}
	\vskip -0.1in
\end{figure}

Given that RepVGG-A0 satisfies \textbf{C1}, we sample a batch of images to depict the difference between the two regarding \textbf{C2} in Fig.~\ref{fig:qarepvgg-a0-before-act}. Our method produces a better activation distribution of smaller standard variances and min/max values, hence ensuring better quantization performance.

\paragraph{Comparison with VGG \cite{simonyan2015a}.} We also product a comparative experiment of vanilla VGG models with the same depth and width. Thus they have the same deployment models. The result is shown in Table~\ref{tab:imaget_net_top1_vgg}. QARepVGG outperforms VGG in view of  both FP32 and INT8 accuracy.
\begin{table}[ht]
	\centering
	\begin{tabular}{lccc}
		\toprule
		Model			&FP32(\%)  & INT8 (\%) \\
		
		\midrule
		RepVGG-A0                              & 72.2& 50.3  \\
		VGG-BN-A0		                      & 70.4& 70.1  \\
		QARepVGG-A0 &72.2&70.4 \\

		
		
		
		
		
		
		\bottomrule
	\end{tabular}
	\caption{Comparsion on ImageNet validation dataset.}
	\label{tab:imaget_net_top1_vgg}
\end{table}

\paragraph{Comparison with OREPA.} OREPA \cite{hu2022online} is also a structural improvement on RepVGG. To reduce the training memory, they introduce linear scaling layers to replace the training-time non-linear norm layers, which could maintain optimization diversity and enhance  representational capacity. We evaluate  the official released  models to obtain its PTQ performance, including OREPA-RepVGG-A0, A1 and A2. As shown in Table~\ref{tab:orepa}, OREPA could achieve an enjoyable FP32 accuracy, while its PTQ results is far from satisfaction. Particularly, OREPA-RepVGG-A2 suffers an accuracy drop of 13.5\% via PTQ.  Therefore, such design cannot address the quantization issue of RepVGG.

\begin{table}[h]
	\setlength{\tabcolsep}{2pt}  
	\centering
	\begin{tabular}{lclc|}
		\toprule
		Variants			&FP32 Acc 	& INT8 Acc	\\
		&   (\%) & (\%) \\
		\midrule
		OREPA-RepVGG-A0 & 73.04& 63.24 (9.80$\downarrow$)\\
		OREPA-RepVGG-A1 & 74.85& 66.35 (8.50$\downarrow$)\\
		OREPA-RepVGG-A2 & 76.72& 63.21 (13.51$\downarrow$)\\
		\bottomrule
	\end{tabular}
	\caption{PTQ performance for OREPA on ImageNet. The structural design of OREPA cannot address the quantization issue of RepVGG.}
	\vskip -0.1in
	\label{tab:orepa}
\end{table}

\section{Proof}
\label{pro:lemma}
\begin{proof}
	We use mathematical induction. Without loss of generality, we use the SGD optimizer. It's not difficult to check the validity of using other optimizers such as Adam \cite{kingma2014adam}.
	
	If $n=0$, then $\vect{\beta_{(3)}^0}=\vect{\beta_{(1)}^0}$ holds, since they both are initialized by \vect{$\mathrm{0}$}.
	
	Suppose $n=k$, $\vect{\beta_{(3)}^k}=\vect{\beta_{(1)}^k}$ holds.
	
	When $n=k+1$,  there is no weight decay for $\vect{\beta}$.
	
	\begin{equation}
		\begin{split}
			\vect{\beta_{(3)}^{k+1}}&=\vect{\beta_{(3)}^k}-lr^{k+1}*\frac{\partial l(\mathrm{W},\vect{\gamma},\vect{\beta})}{\partial \vect{\beta_{(3)}^k}}\\
			&=\vect{\beta_{(3)}^k}-lr^{k+1}*\frac{\partial l(\mathrm{W},\vect{\gamma},\vect{\beta})}{\partial  \vect{\mathrm{M}_{(2)}^{k+1}}}\\
			&=\vect{\beta_{(1)}^k}-lr^{k+1}*\frac{\partial l(\mathrm{W},\vect{\gamma},\vect{\beta})}{\partial \vect{\mathrm{M}_{(2)}^{k+1}}}=\vect{\beta_{(1)}^{k+1}}.
		\end{split}
	\end{equation}
	
	Q.E.D.
\end{proof}

\section{Figures}
\begin{figure}[ht]
	\centering
	\includegraphics[width=\columnwidth]{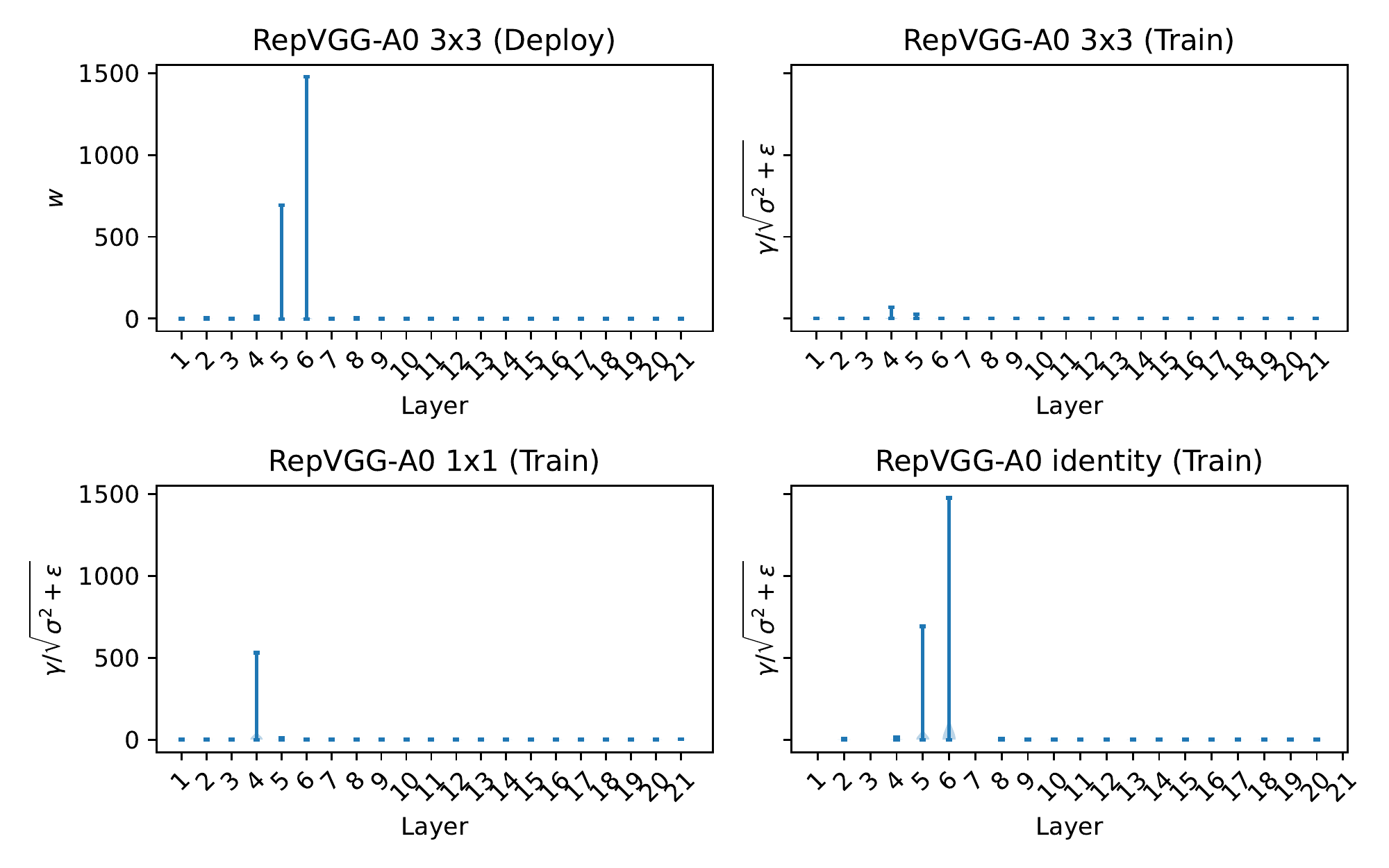}
	\caption{Violin plot of convolutional weights in each layer of RepVGG-A0  trained without custom $L_2$ (\textbf{S1}). The weight distributions of layer 5 and 6 under the deploy setting have large variances, incurring large quantization errors (\textbf{C1 violation}).} 
	\label{fig:repvgg-a0-nocwd-w-violin}
	\vskip -0.1in
\end{figure}

\subsection{Weight Distribution}


\begin{figure}[ht]
	\centering
	\includegraphics[width=0.9\columnwidth]{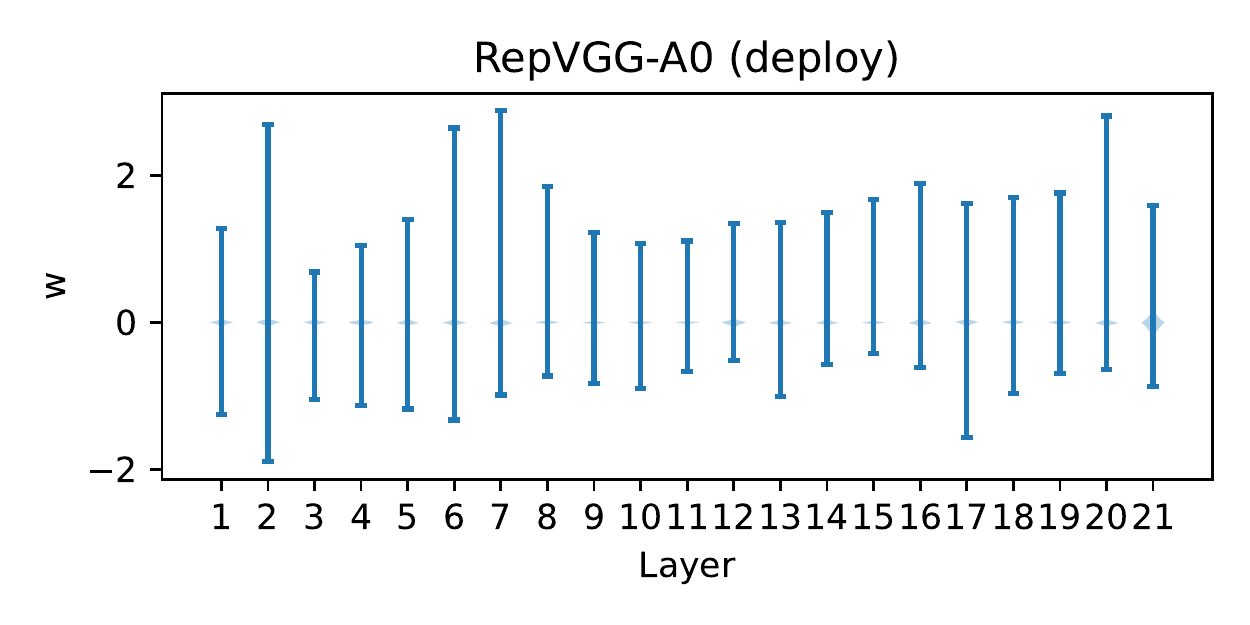}
	\caption{Violin plot of RepVGG-A0 convolutional weights when trained removing $\gamma/\sqrt{\sigma^2+\epsilon}$  in $L_2$.}
	\label{fig:qarepvgg-a0-boxplot}
\end{figure}

\begin{figure*}[ht]
	\centering
	\begin{subfigure}{\linewidth}
		\includegraphics[width=\textwidth]{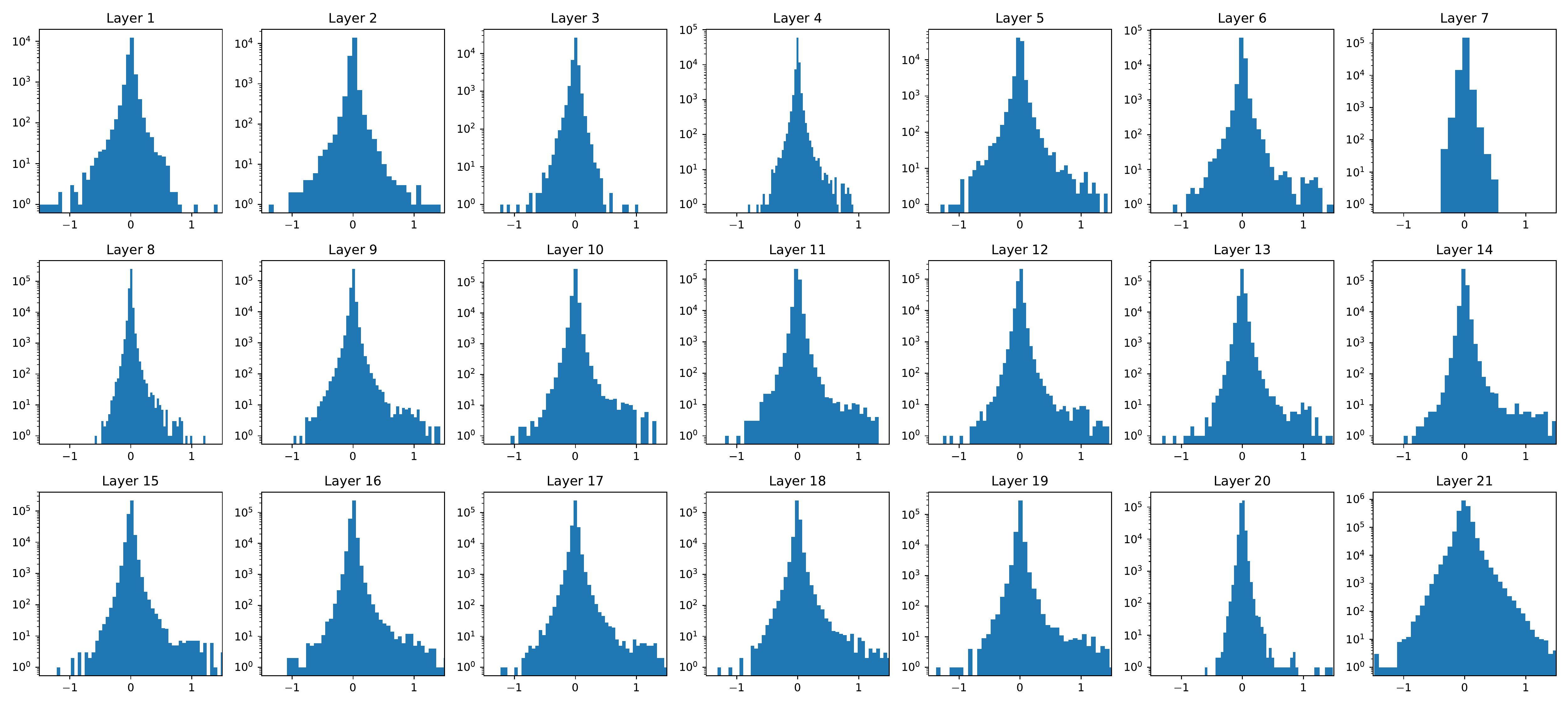}
		\caption{RepVGG-A0 trained with the default custom $L_2$}
		\label{fig:repvgg-a0-weight-dist}
	\end{subfigure}
	\centering
	\begin{subfigure}{\linewidth}
		\includegraphics[width=\textwidth]{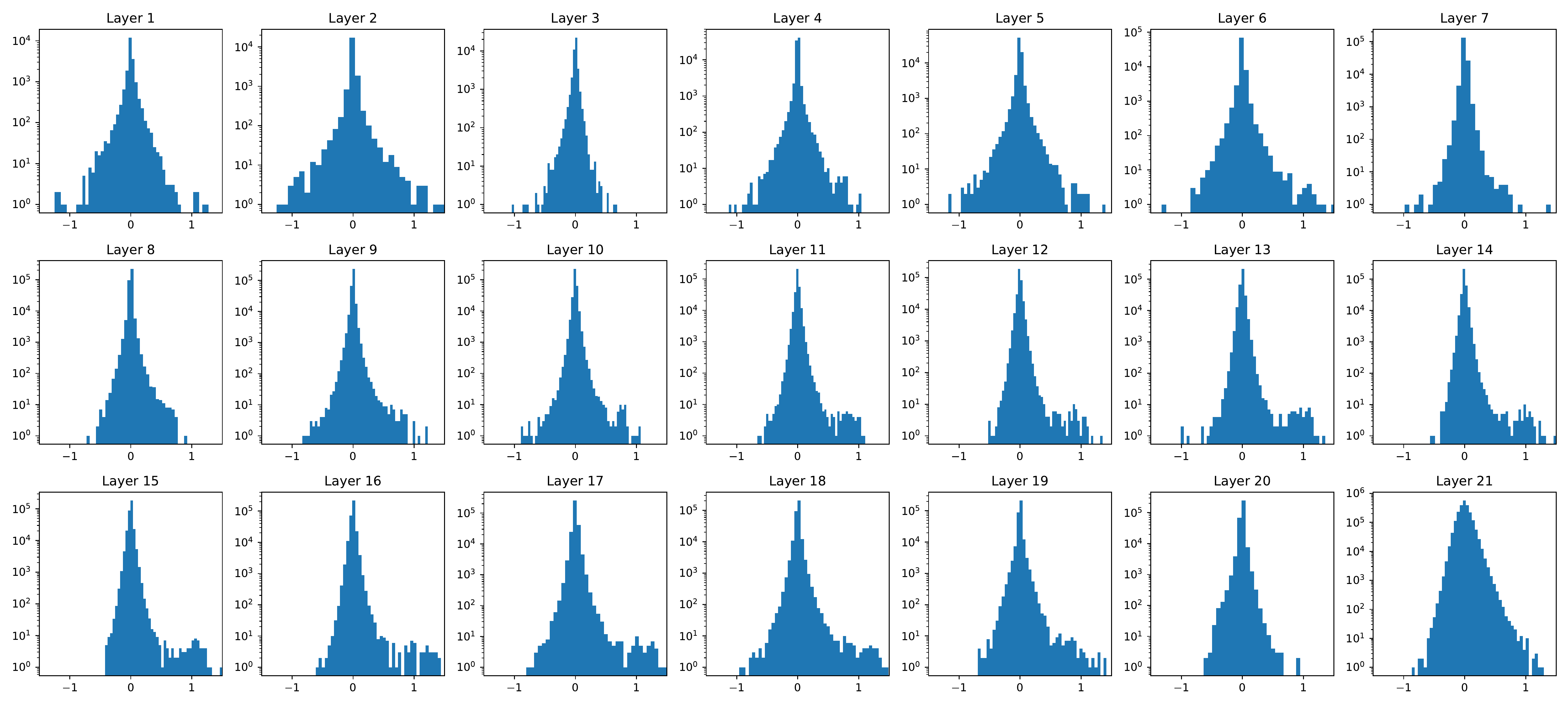}
		\caption{RepVGG-A0 trained removing $\gamma/\sqrt{\sigma^2+\epsilon}$ in custom $L_2$}
		\label{fig:qarepvgg-a0-weight-dist}
	\end{subfigure}
	\caption{Fused weight distribution of RepVGG-A0 trained under different $L_2$ settings.}
	\label{fig:weight-dist}
\end{figure*}

Due to built-in weight normalization, either custom L2 or vanilla L2, the convolutional weights generally center around 0 and span unbiasedly for RepVGG-A0, see Figure~\ref{fig:weight-dist}. We also show the violin plot of RepVGG-A0 trained removing the coefficient $\gamma/\sqrt{\sigma^2+\epsilon}$ in $L_2$ in Figure~\ref{fig:qarepvgg-a0-boxplot}. These figures show one of the priors is satisfied to be quantization-friendly.

\begin{figure}[ht]
	\centering
	\includegraphics[width=0.9\columnwidth]{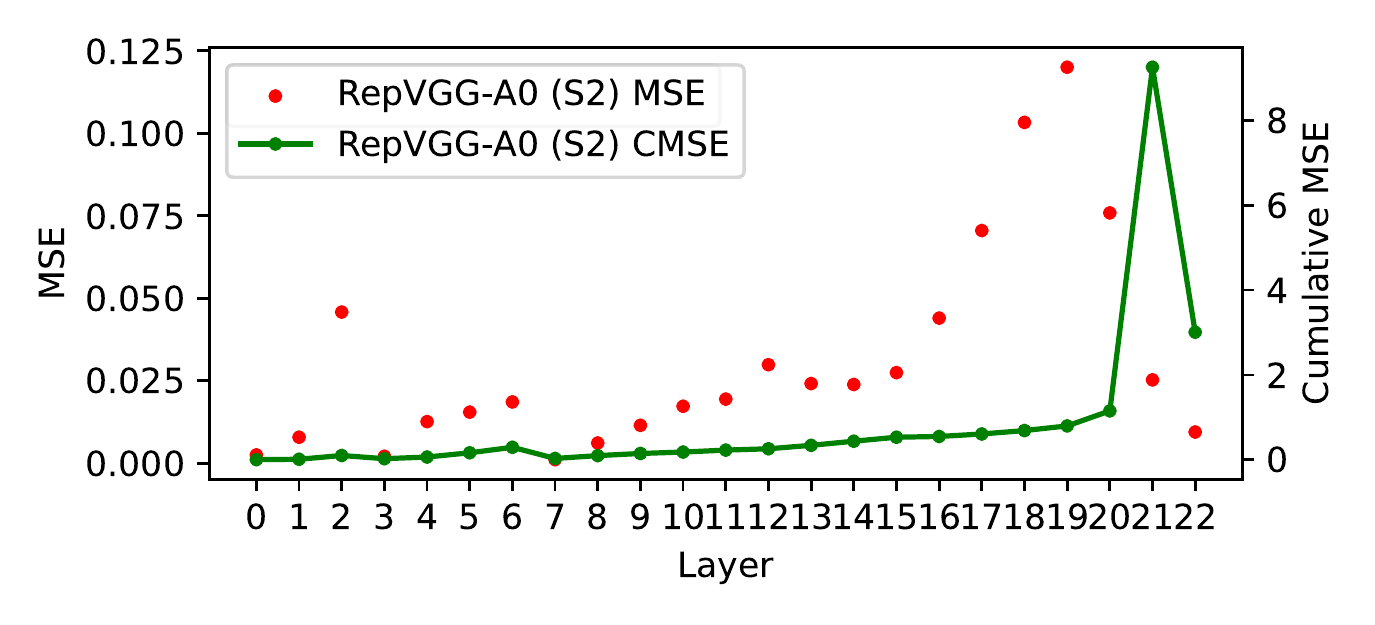}
	\caption{MSE and Cumulative MSE of RepVGG-A0 (\textbf{S2}).} 
	\label{fig:repvgg-a0-v6-s2}
\end{figure}

\begin{figure}[ht]
	\centering
	\includegraphics[width=0.9\columnwidth]{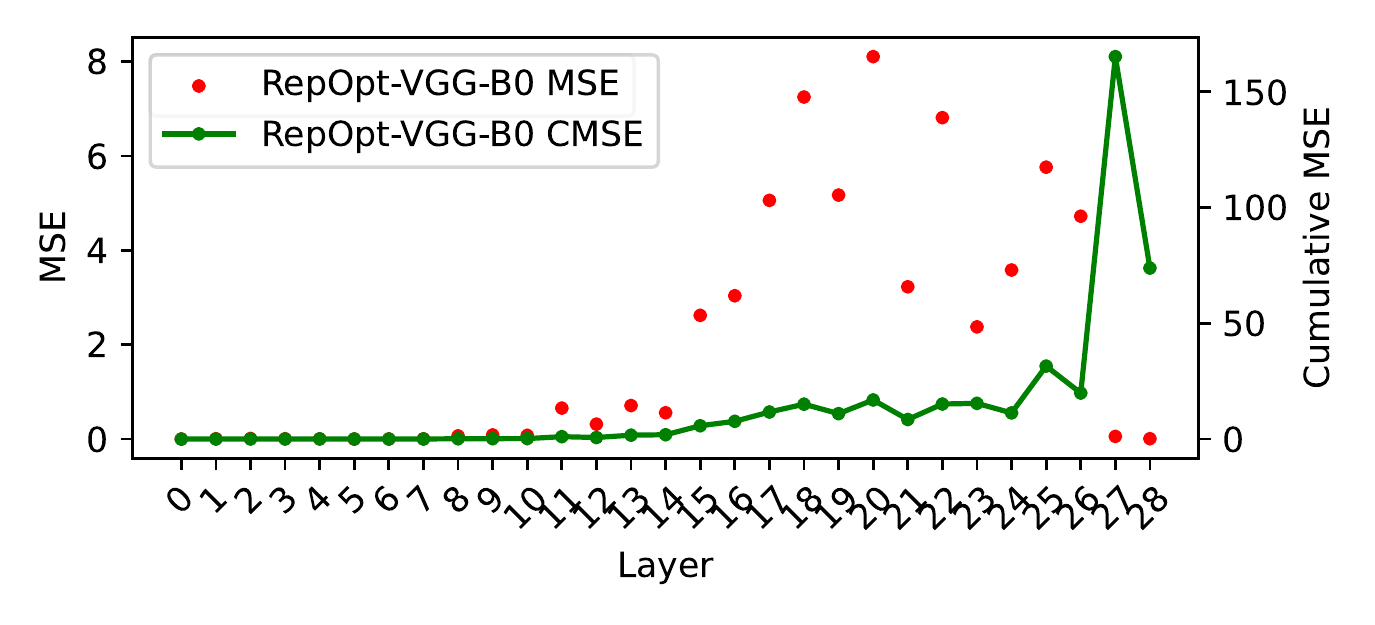}
	\caption{MSE and Cumulative MSE of RepOpt-VGG-B0.} 
	\label{fig:repopt-vgg-b0-mse}
\end{figure}

\begin{figure}[ht]
	\centering
	\begin{subfigure}{\columnwidth}
		\centering
		\includegraphics[width=0.9\columnwidth]{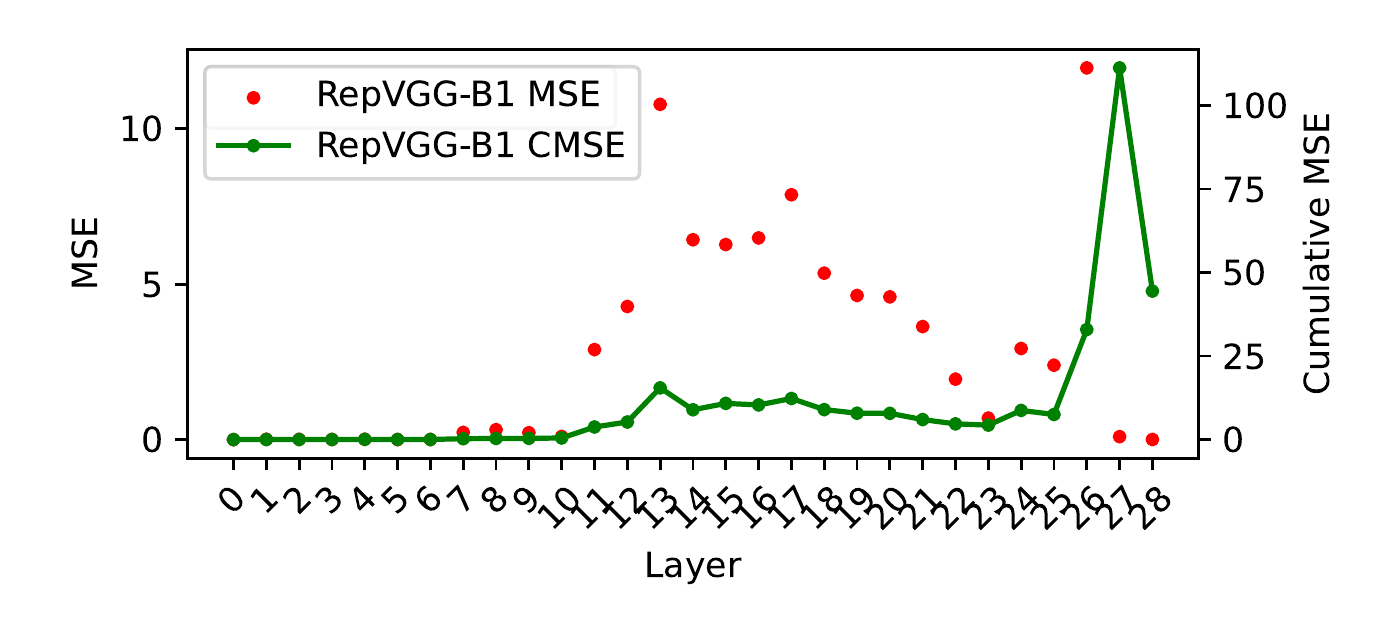}
		\caption{RepVGG-B1}
	\end{subfigure}
	\begin{subfigure}{\columnwidth}
		\centering
		\includegraphics[width=0.9\columnwidth]{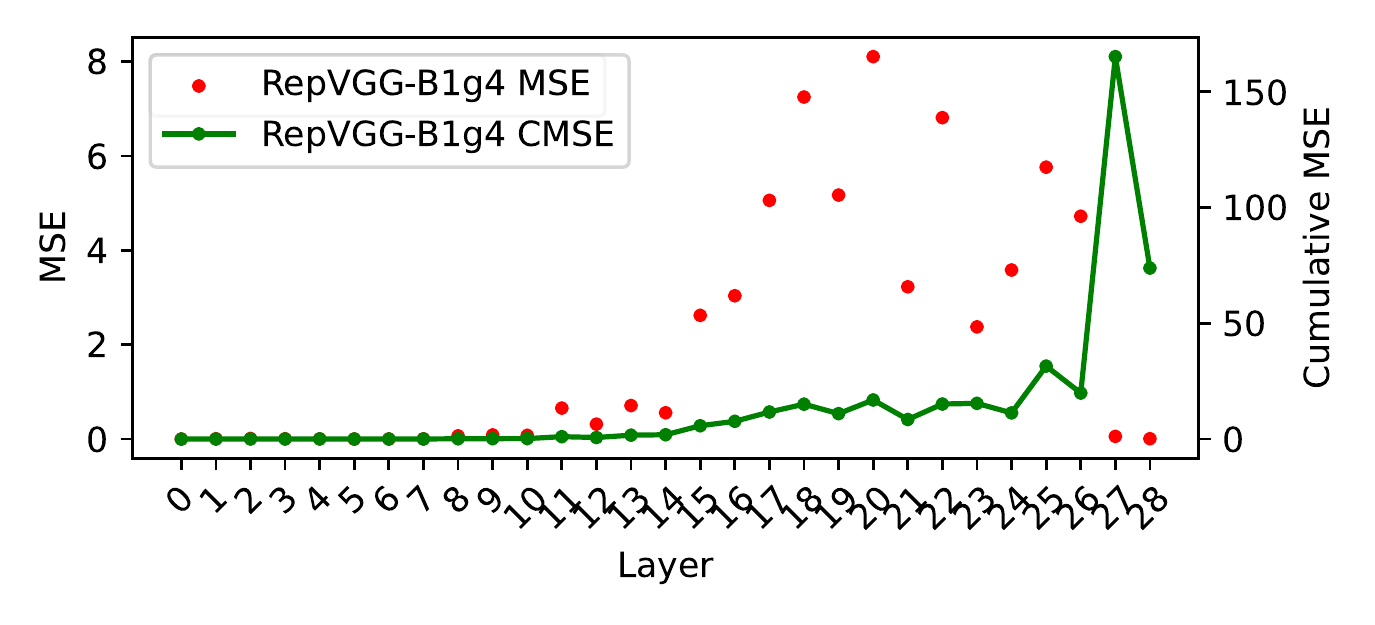}
		\caption{RepVGG-B1g4}
	\end{subfigure}
	\caption{MSE and Cumulative MSE of RepVGG-B1 and RepVGG-B1g4.} 
	\label{fig:repvgg-b1-mse}
\end{figure}

\begin{figure}[ht]
	\centering
	\includegraphics[width=0.9\columnwidth]{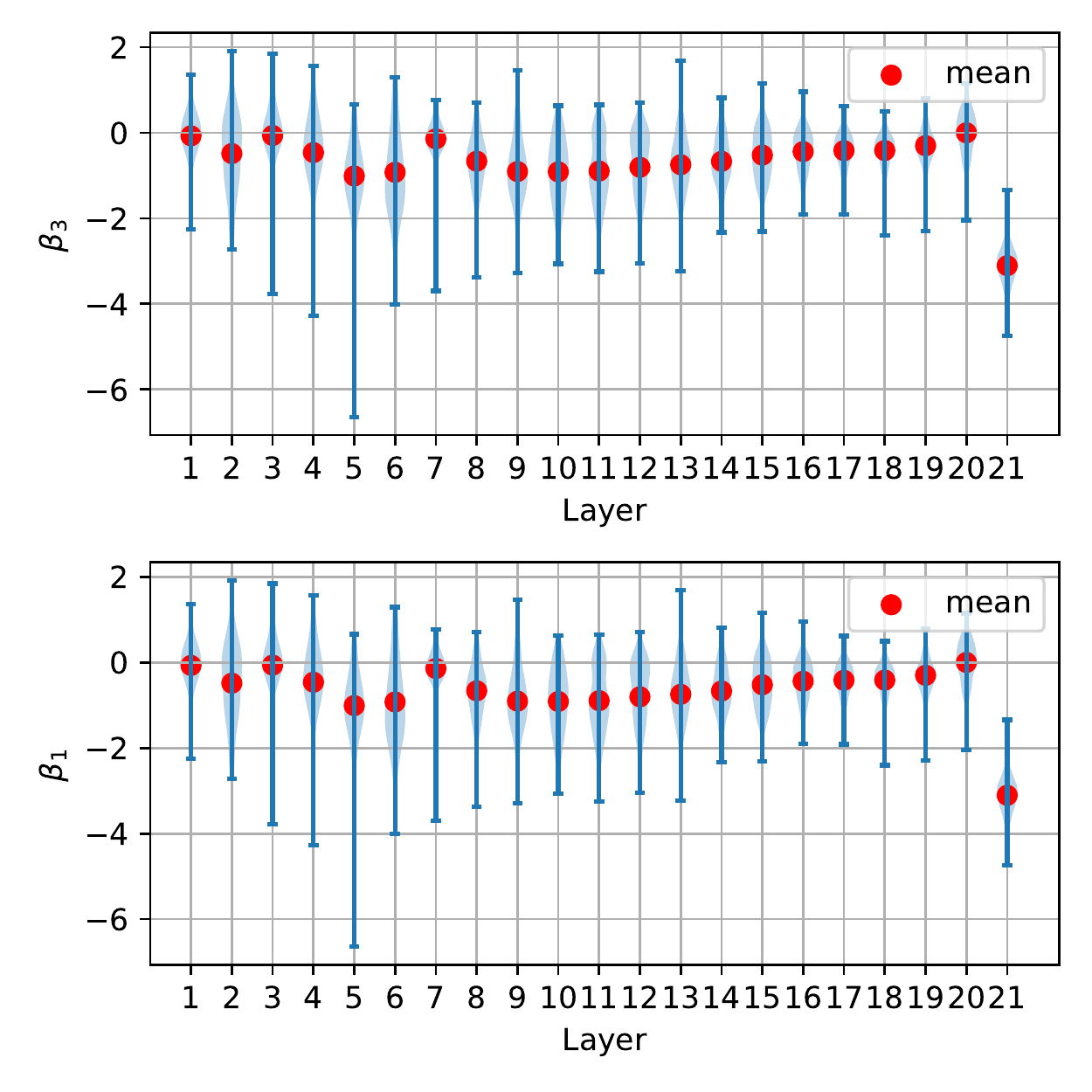}
	\caption{BN bias distribution of RepVGG-A0 convolutional weights trained in \textbf{S2}. $\beta_i$ for BN after Conv $i \times i$.} 
	\label{fig:repvgg-a0-v6-s2-bias}
\end{figure}

\begin{figure*}[ht]
	\centering
	\includegraphics[width=0.9\textwidth]{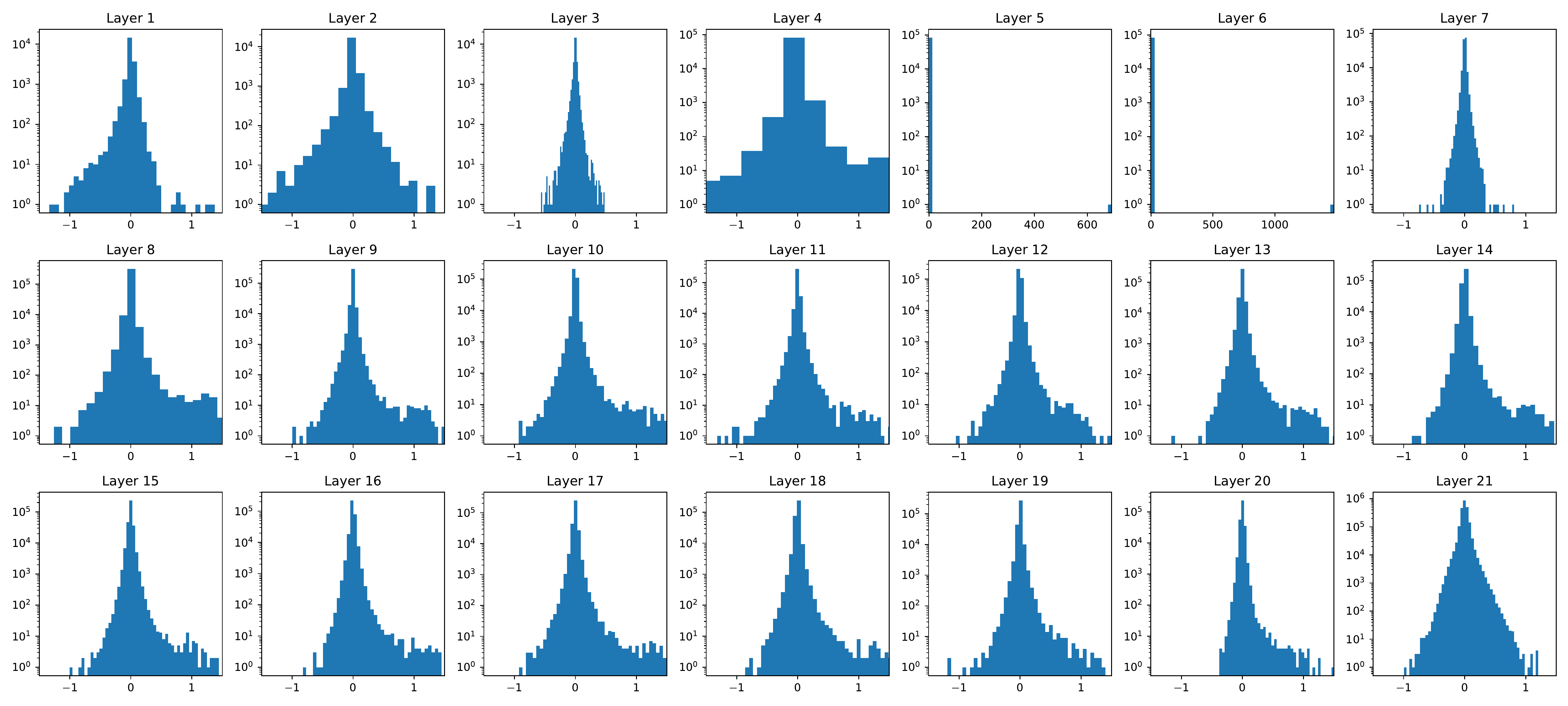}
	\caption{Distribution of RepVGG-A0 fused convolutional weights trained without custom $L_2$ regularization (\textbf{S1}). Layer 5 and 6 becomes intractable and violate \textbf{C1}.}
	\label{fig:repvgg-a0-nocwd-w-dist}
\end{figure*}



\begin{figure*}[ht]
	\centering
	\begin{subfigure}{\linewidth}
		\includegraphics[width=\textwidth]{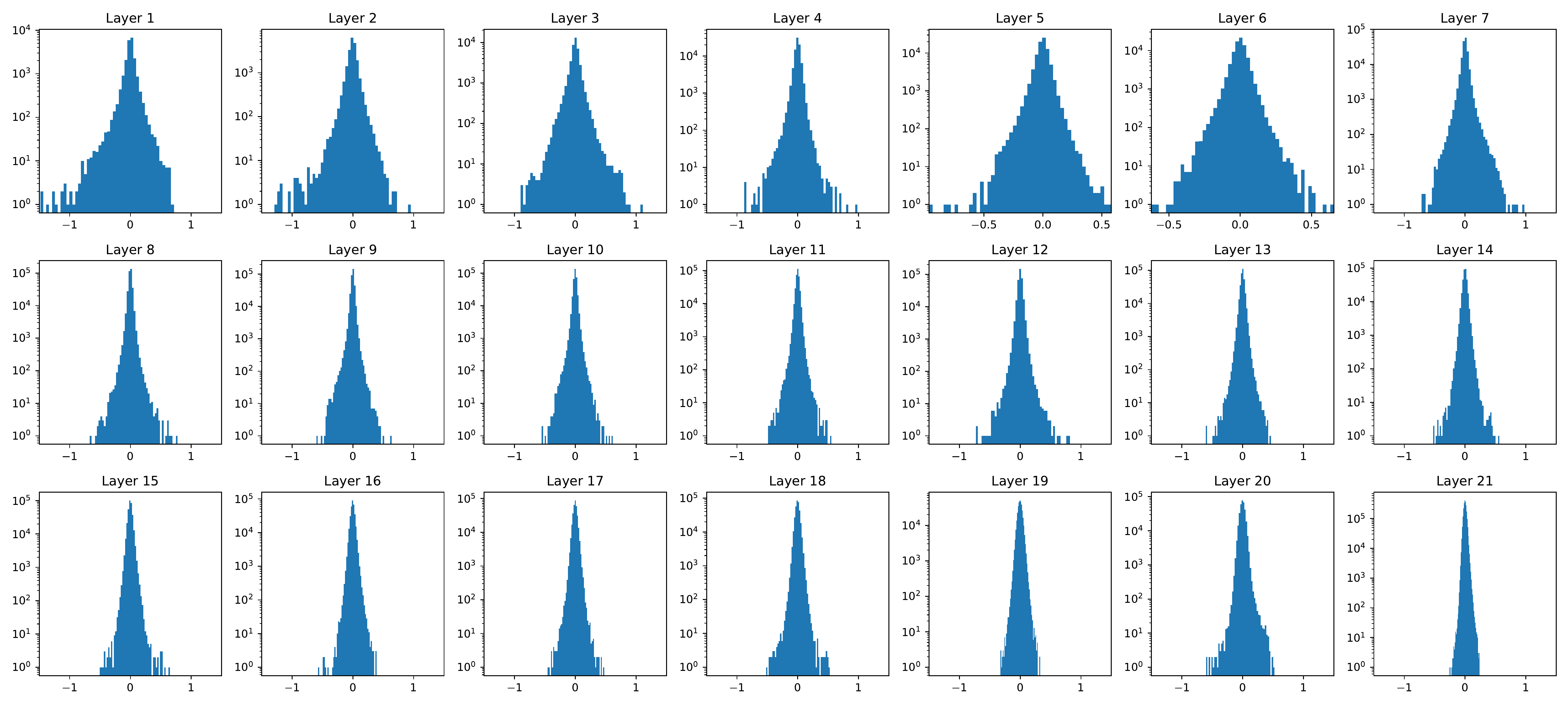}
		\caption{RepVGG-A0 Conv $3\times3$}
		\label{fig:repvgg-a0-rbr_dense-weight-dist}
	\end{subfigure}
	\centering
	\begin{subfigure}{\linewidth}
		\includegraphics[width=\textwidth]{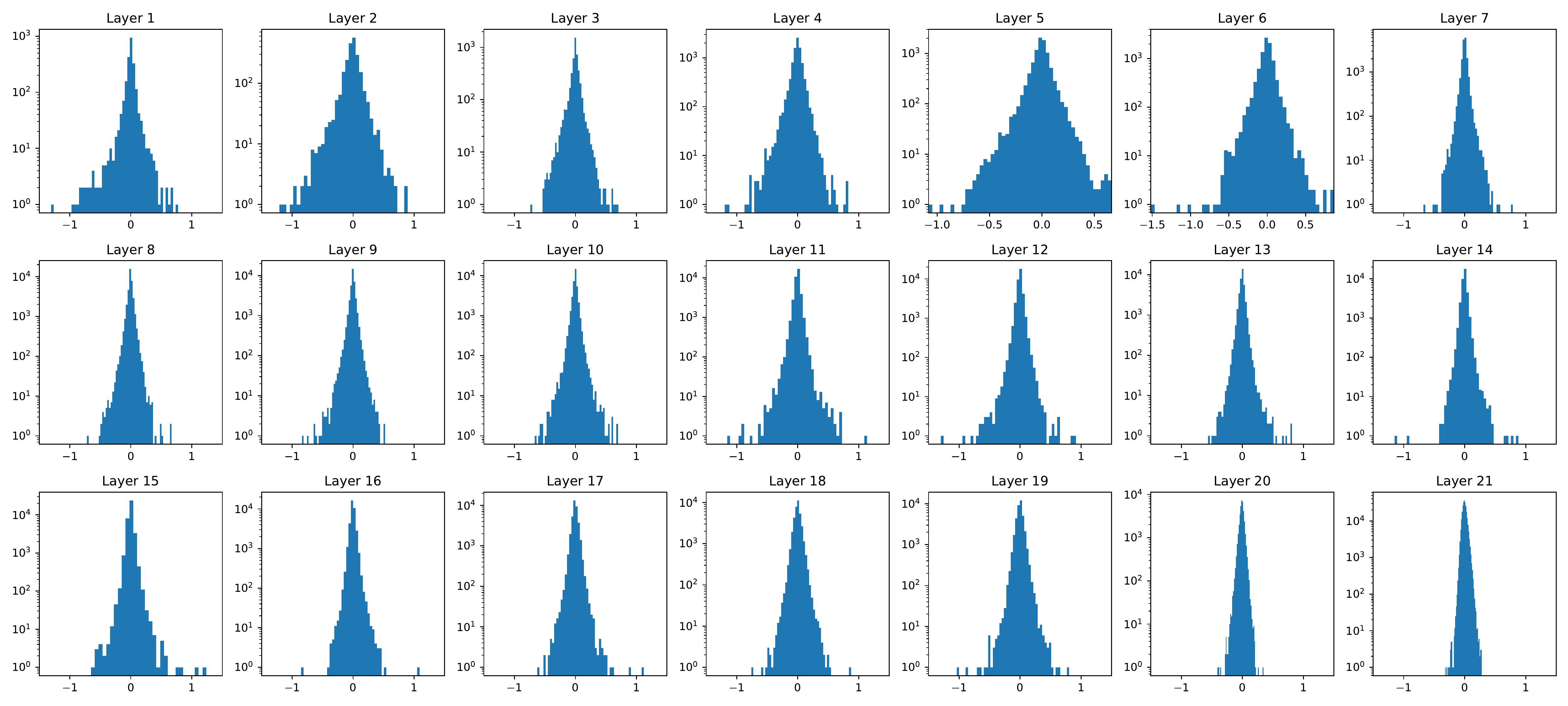}
		\caption{RepVGG-A0 Conv $1\times1$}
		\label{fig:qarepvgg-a0-rbr_1x1-weight-dist}
	\end{subfigure}
	\caption{Weight distribution in each layer of RepVGG-A0 trained without custom $L_2$ regularization.}
	\label{fig:weight-dist-no-cwd-l2}
\end{figure*}

\begin{figure*}[ht]
	\centering
	\begin{subfigure}{\linewidth}
		\includegraphics[width=\textwidth]{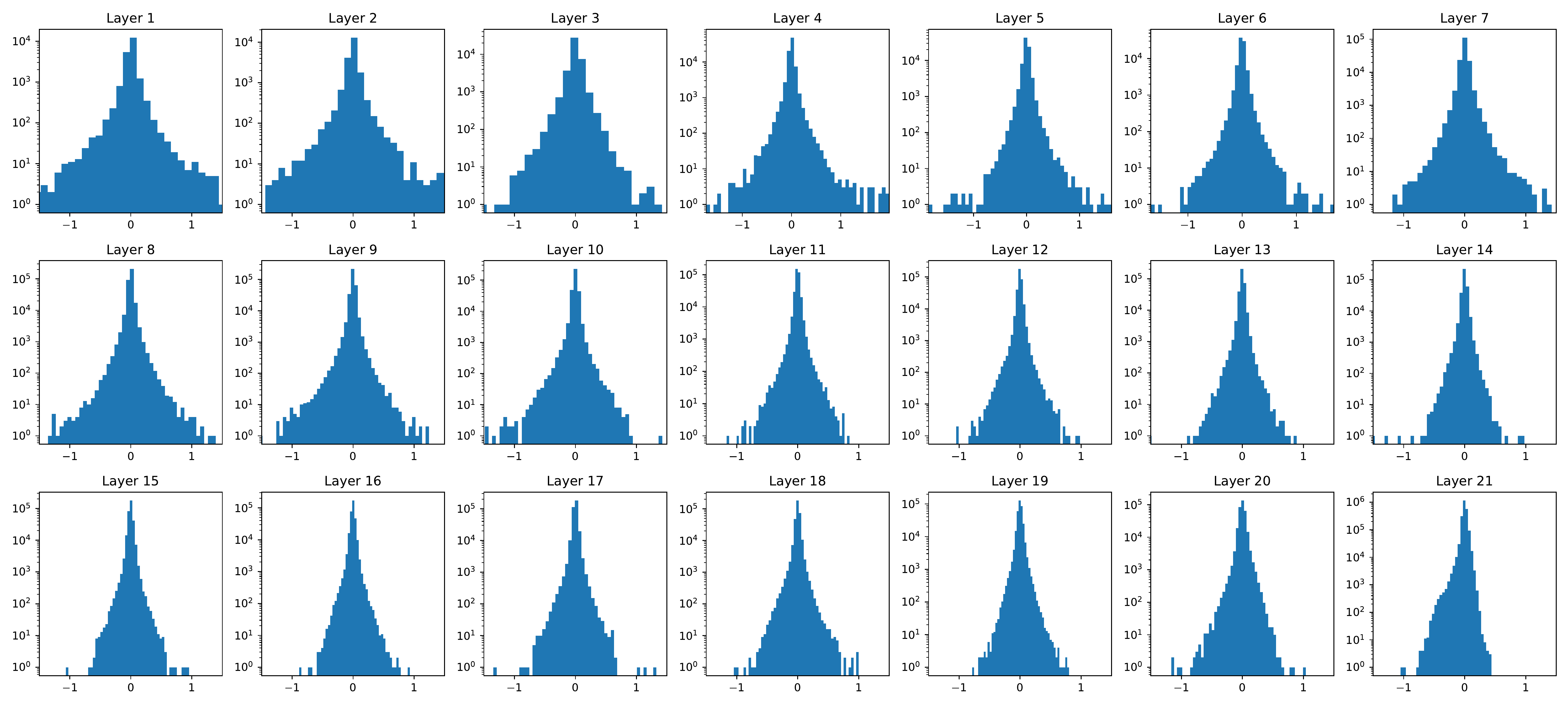}
		\caption{RepVGG-A0 Conv $3\times3$}
		\label{fig:repvgg-a0-rbr_dense-weight-dist}
	\end{subfigure}
	\centering
	\begin{subfigure}{\linewidth}
		\includegraphics[width=\textwidth]{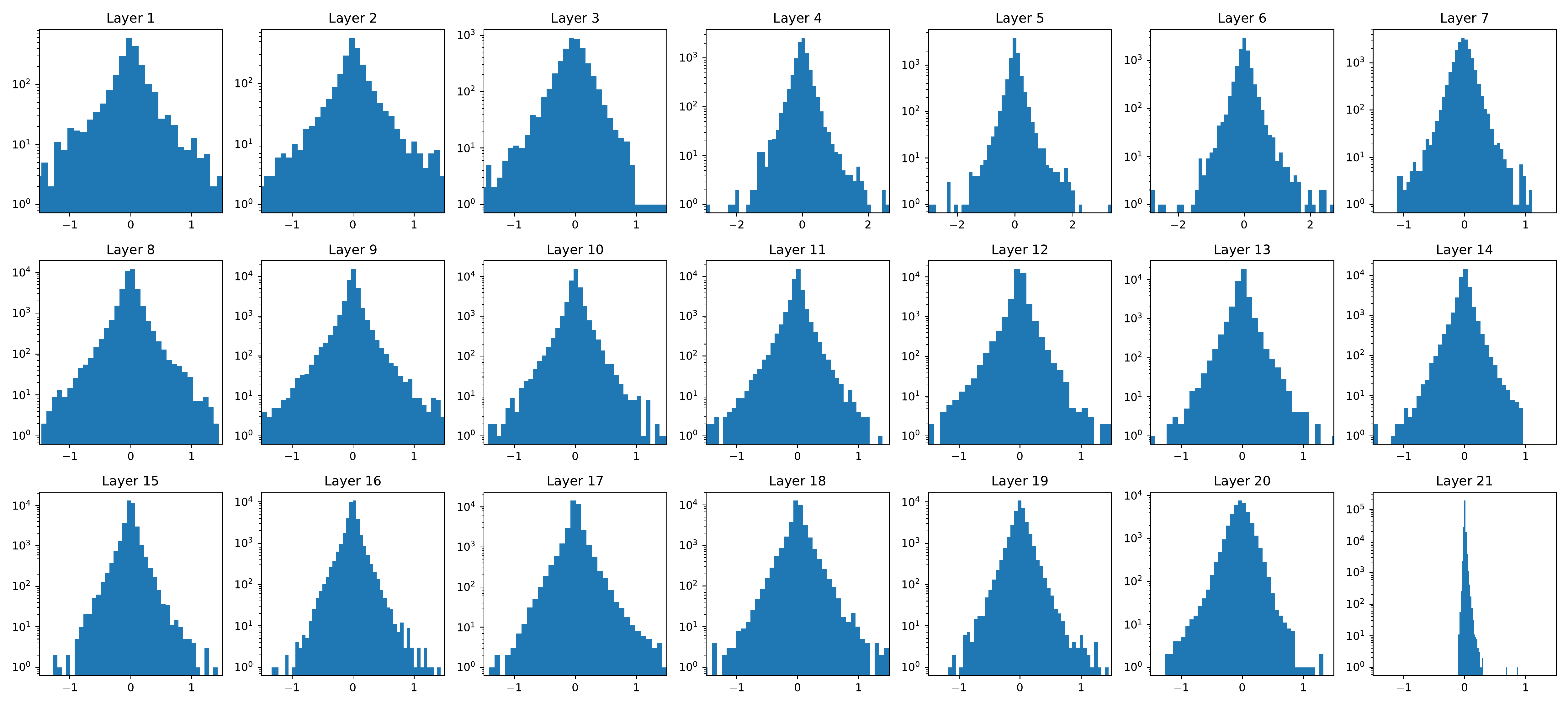}
		\caption{RepVGG-A0 Conv $1\times1$}
		\label{fig:repvgg-a0-rbr_1x1-weight-dist}
	\end{subfigure}
	\caption{Weight distribution in each layer of RepVGG-A0 trained when removing $\gamma/\sqrt{\sigma^2+\epsilon}$ in custom $L_2$ regularization.}
	\label{fig:weight-dist-no-coeff-cwd-l2}
\end{figure*}

\begin{figure*}[ht]
	\begin{subfigure}{\linewidth}
		\centering
		\includegraphics[width=0.9\textwidth]{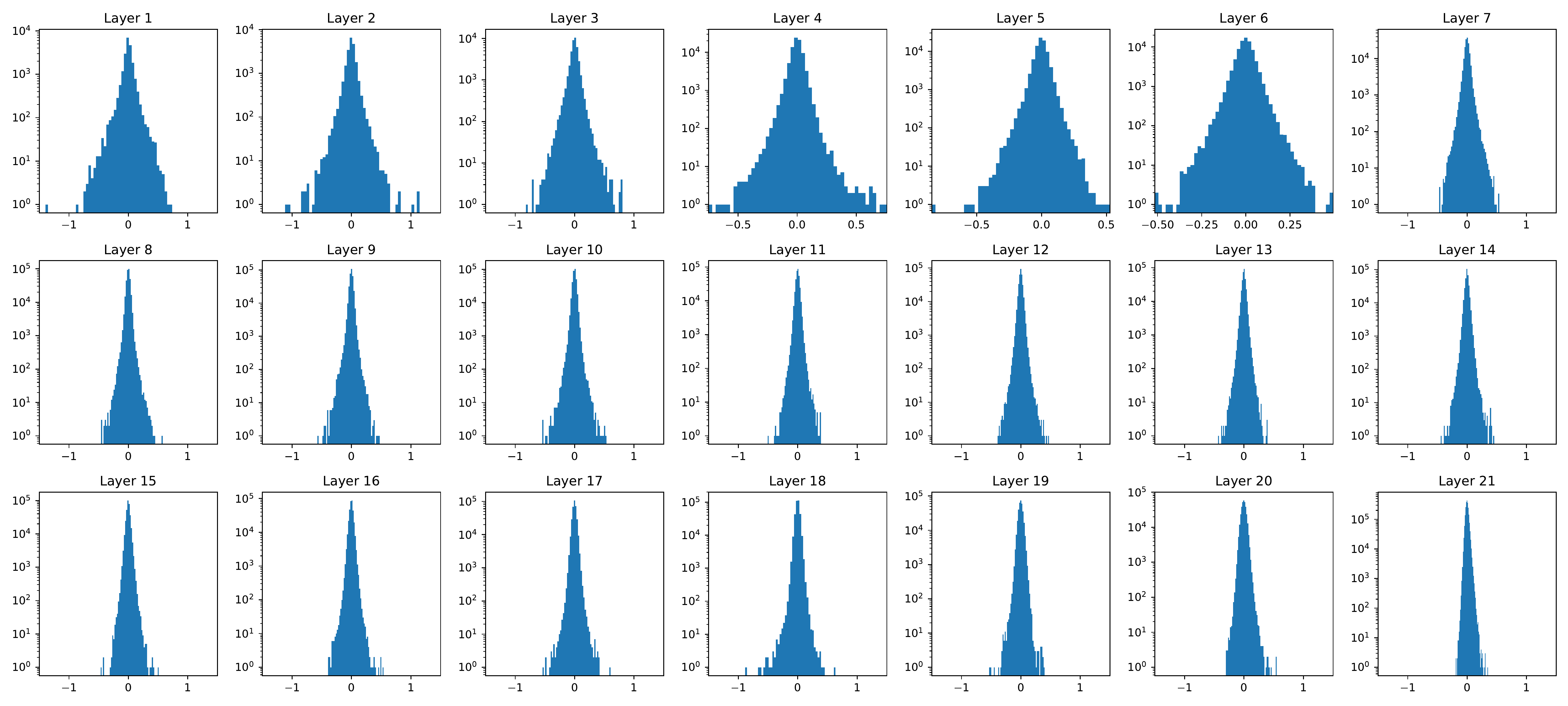}
		\caption{QARepVGG-A0 Conv $3\times3$}
		\label{fig:qarepvgg-a0-rbr_dense-weight-dist}
	\end{subfigure}
	\begin{subfigure}{\linewidth}  \centering
		\includegraphics[width=0.9\textwidth]{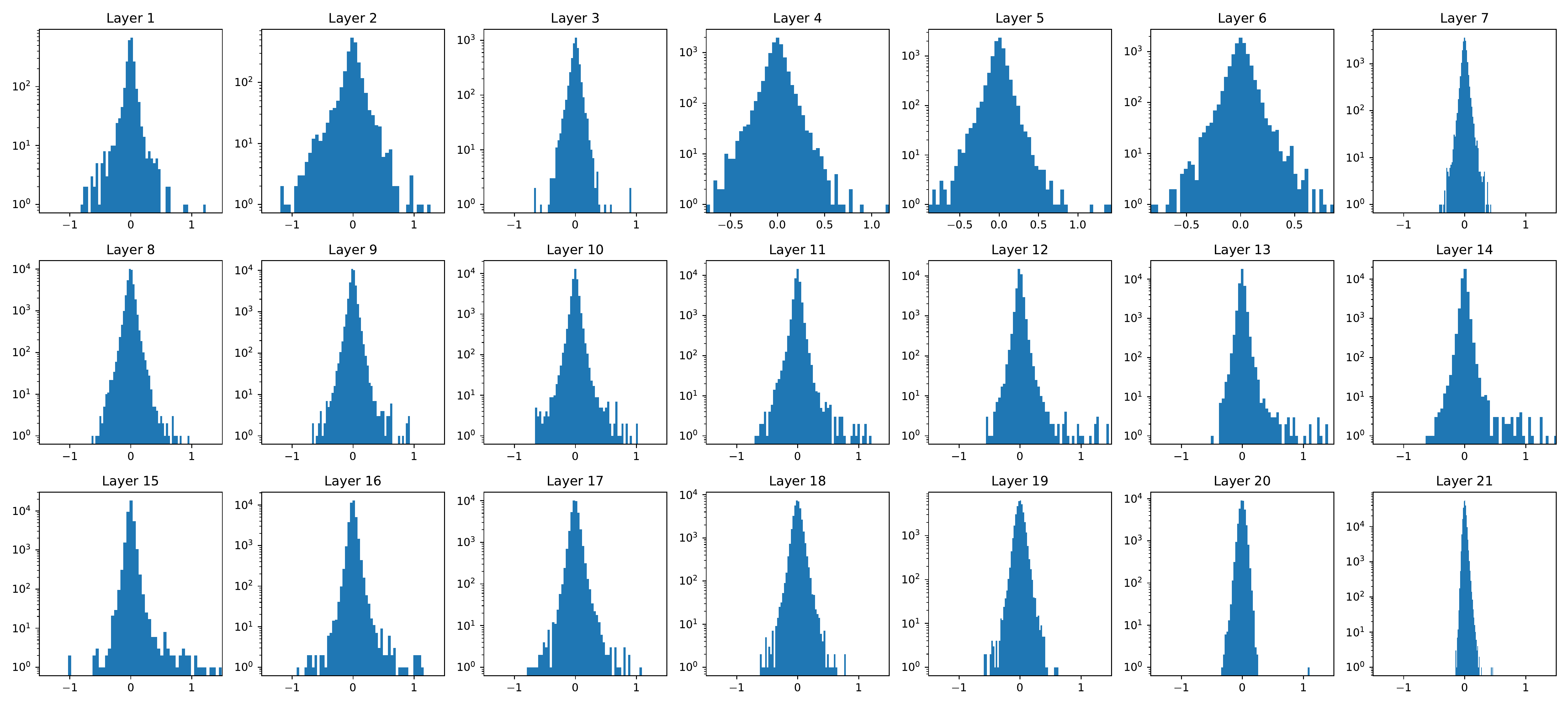}
		\caption{QARepVGG-A0 Conv $1\times1$}
		\label{fig:qarepvgg-a0-rbr_1x1-weight-dist}
	\end{subfigure}
	\begin{subfigure}{\linewidth}  \centering
		\includegraphics[width=0.9\textwidth]{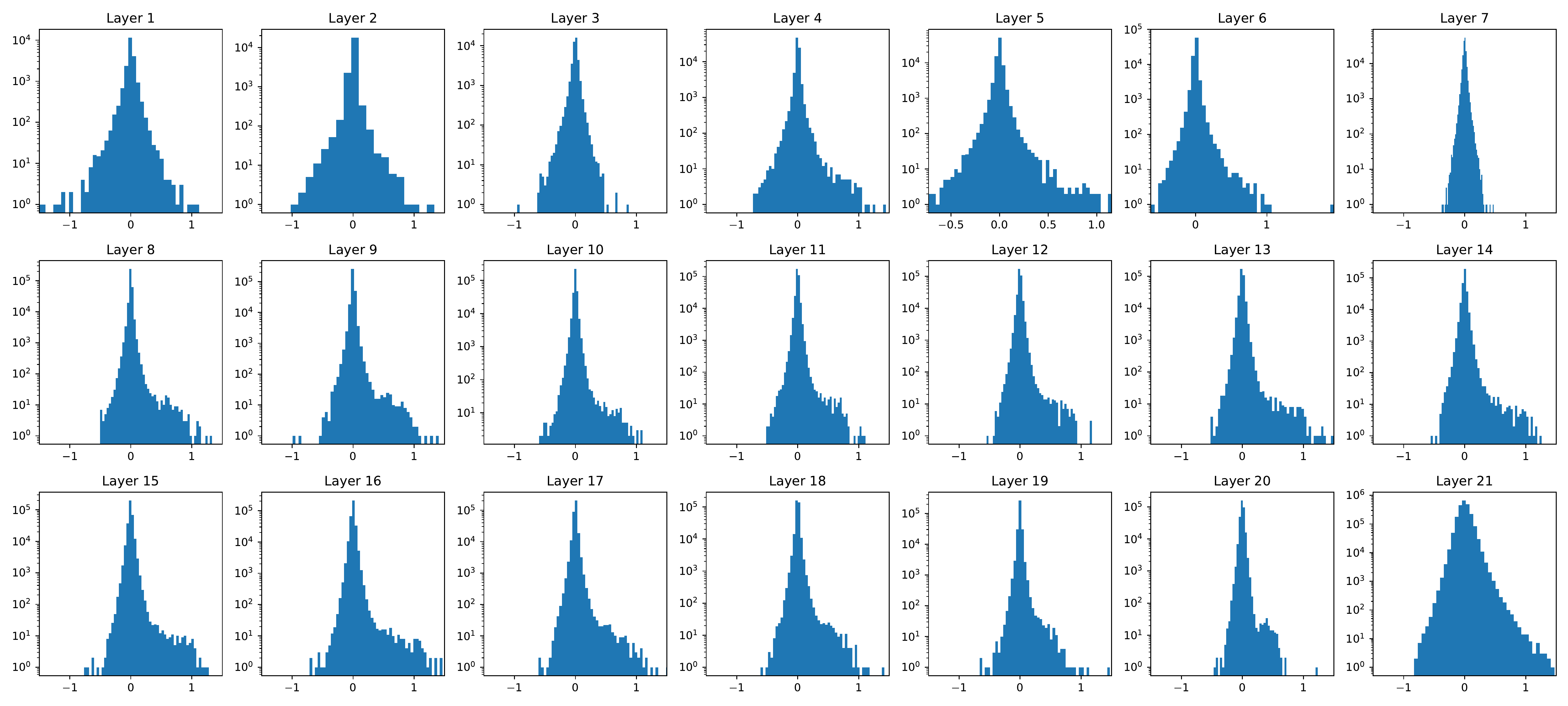}
		\caption{Fused QARepVGG-A0}
		\label{fig:qarepvgg-a0-rbr_fused-weight-dist}
	\end{subfigure}
	\caption{Weight distribution of Conv $3\times3$ and $1\times1$ and fused Conv in each layer of QARepVGG-A0.}
	\label{fig:qa-weight-dist-no-cwd-l2}
\end{figure*}
\end{document}